%% file: main.tex
\documentclass{article}
\usepackage[main,final]{neurips_2026}
\input{tex/preamble/packages.tex}
\makeatletter
\renewcommand{\@notice}{}
\makeatother

\title{Visual Agentic Memory: Enabling Online Long Video Understanding via Online Indexing, Hierarchical Memory, and Agentic Retrieval}
\author{
  Aiden Yiliu Li$^{1}$ \quad
  Nels Numan$^{1}$ \quad
  Anthony Steed$^{1}$ \\
  $^{1}$University College London \\
  \texttt{yiliu.li@outlook}, \texttt{a.steed@ucl.ac.uk}
}
\date{}

\begin{document}

\maketitle

\begin{abstract}
\input{tex/frontmatter/abstract}
\end{abstract}

\input{sections/introduction}
\input{sections/related_work}
\input{sections/methodology}
\input{sections/evaluation}
\input{sections/conclusion}

\bibliographystyle{plainnat}
\bibliography{references.bib}

\appendix
\input{sections/appendix}

\end{document}

%% file: tex/preamble/packages.tex
\usepackage[T1]{fontenc}
\usepackage[utf8]{inputenc}
\usepackage[english]{babel}
\usepackage{url}
\usepackage{siunitx}
\usepackage{graphicx}
\usepackage{graphics} % for pdf, bitmapped graphics files
\usepackage{amsmath}
\usepackage{latexsym}
\usepackage{amscd}% for commutative diagrams
\usepackage{mathrsfs} %this package is for the script font \mathscr
\usepackage{relsize}
\usepackage{delarray}

\usepackage{graphicx} % for youtube image link

\usepackage{xcolor} % highlight text 
\providecommand{\newglossaryentry}[2]{}

\usepackage{theorem}
\usepackage{changepage}
\usepackage{euscript}
\usepackage{textcomp}
\usepackage{esvect}
\usepackage{parskip}
\usepackage{placeins}
\usepackage{subcaption}
\usepackage{array}
\usepackage{tabularx}
\usepackage{delarray}
\usepackage{stmaryrd}
\usepackage{graphpap}
\usepackage{makeidx}
\usepackage{enumerate}
\usepackage{esint}
\usepackage{caption}
\usepackage{smartdiagram}
\usesmartdiagramlibrary{additions}

\usepackage{color}
\definecolor{igreen}{rgb}{0,0,0}
\usepackage{xcolor, colortbl}
\colorlet{gred}{black}
\colorlet{mamber}{black}
\colorlet{grown}{black}
\colorlet{bled}{black}
\colorlet{waters}{black} % Define color for the water
\colorlet{water}{black} % Define color for the water
\definecolor{grin}{HTML}{000000}
\usepackage{rotating}

\usepackage{arydshln}
\usepackage{booktabs}
\usepackage{graphicx}
\usepackage{tikz}
\usepackage{tikz-3dplot}
\usetikzlibrary{
arrows,
arrows.meta,
automata,
backgrounds,
calc,
decorations,
decorations.pathmorphing,
decorations.pathreplacing,
decorations.fractals,
external,
fit,
matrix,
petri,
positioning,
shadows,
shapes,
shapes.multipart,
topaths,
intersections
}
\usepackage{eso-pic}
\def\ba{\begin{array}}
\def\ea{\end{array}}
\def\beann{\begin{eqnarray*}}
\def\eeann{\end{eqnarray*}}
\def\bea{\begin{eqnarray}}
\def\eea{\end{eqnarray}}

\makeatletter
\newlength\qvec@height
\newlength\qvec@depth
\newlength\qvec@width
\newcommand{\qvec}[2][]{
    \settoheight{\qvec@height}{$#2$}
    \settodepth{\qvec@depth}{$#2$}
    \settowidth{\qvec@width}{$#2$}
  \def\qvec@arg{#1}
  \raisebox{.2ex}{\raisebox{\qvec@height}{\rlap{% 
    \kern.05em
    \begin{tikzpicture}[scale=1,shorten >=-3pt,shorten <=-3pt]
    \pgfsetroundcap
    \coordinate (Stx) at (.05em,0) ;
		\coordinate (Arx) at (\qvec@width-.05em,0) ;
    \draw[->](Stx) to[bend left] (Arx);
    \end{tikzpicture}
  }}}
  #2
}
\makeatother
\makeatletter
\newlength\pvec@height
\newlength\pvec@depth
\newlength\pvec@width
\newcommand{\pvec}[2][]{
    \settoheight{\pvec@height}{$#2$}
    \settodepth{\pvec@depth}{$#2$}
    \settowidth{\pvec@width}{$#2$}
  \def\pvec@arg{#1}
  \raisebox{.2ex}{\raisebox{\pvec@height}{\rlap{% 
    \kern.05em
    \begin{tikzpicture}[scale=1,shorten >=-3pt,shorten <=-3pt]
    \pgfsetroundcap
    \coordinate (Stx) at (.05em,0) ;
		\coordinate (Arx) at (\pvec@width-.05em,0) ;
    \draw[->](Stx) to[bend right] (Arx);
    \end{tikzpicture}
  }}}
  #2
}
\makeatother
\makeatletter
\newlength\vvec@height%
\newlength\vvec@depth%
\newlength\vvec@width%
\newcommand{\vvec}[2][]{%
  \ifmmode%
    \settoheight{\vvec@height}{$#2$}%
    \settodepth{\vvec@depth}{$#2$}%
    \settowidth{\vvec@width}{$#2$}%
  \else 
    \settoheight{\vvec@height}{#2}%
    \settodepth{\vvec@depth}{#2}%
    \settowidth{\vvec@width}{#2}%
  \fi%
  \def\vvec@arg{#1}%
  \def\vvec@dd{:}%
  \def\vvec@d{.}%
  \raisebox{.2ex}{\raisebox{\vvec@height}{\rlap{%
    \kern.05em%
    \begin{tikzpicture}[scale=1]
    \pgfsetroundcap
    \draw (.05em,0)--(\vvec@width-.05em,0);
    \draw (\vvec@width-.05em,0)--(\vvec@width-.15em, .075em);
    \draw (\vvec@width-.05em,0)--(\vvec@width-.15em,-.075em);
    \ifx\vvec@arg\vvec@d%
      \fill(\vvec@width*.45,.5ex) circle (.5pt);%
    \else\ifx\vvec@arg\vvec@dd%
      \fill(\vvec@width*.30,.5ex) circle (.5pt);%
      \fill(\vvec@width*.65,.5ex) circle (.5pt);%
    \fi\fi%
    \end{tikzpicture}%
  }}}%
  #2%
}
\makeatother
\def\ba{\begin{array}}
\def\ea{\end{array}}
\def\beann{\begin{eqnarray*}}
\def\eeann{\end{eqnarray*}}
\def\bea{\begin{eqnarray}}
\def\eea{\end{eqnarray}}

\usepackage{multirow}
\usepackage{hyperref}
\usepackage{lipsum}
\usepackage{tikz-cd}
\usepackage{float}

\usepackage{xcolor}
\usepackage{listings}
\usepackage{xcolor}

\definecolor{codegreen}{rgb}{0,0,0}
\definecolor{codegray}{rgb}{0,0,0}
\definecolor{codepurple}{rgb}{0,0,0}
\definecolor{backcolour}{rgb}{1,1,1}

\lstdefinestyle{mystyle}{
  backgroundcolor=\color{backcolour}, commentstyle=\color{codegreen},
  keywordstyle=\color{black},
  numberstyle=\tiny\color{codegray},
  stringstyle=\color{codepurple},
  basicstyle=\ttfamily\footnotesize,
  breakatwhitespace=false,         
  breaklines=true,                 
  captionpos=b,                    
  keepspaces=true,                 
  numbers=left,                    
  numbersep=5pt,                  
  showspaces=false,                
  showstringspaces=false,
  showtabs=false,                  
  tabsize=2
}

\usepackage{amsfonts}

\usepackage{algorithm}
\usepackage{algpseudocode}

\usepackage{pgfgantt}

\usepackage{makecell} %multiline in tables

\usepackage{standalone}

\usepackage{lscape}

%% file: tex/frontmatter/abstract.tex
Long video understanding requires more than large context windows. It also needs a memory mechanism that decides what visual evidence to retain, keeps it searchable over long horizons, and grounds later reasoning in recoverable observations rather than compressed latent state alone. We propose Visual Agentic Memory (VAM), a training-free framework with three components. \textit{Online Indexing} supports selective evidence retention under streaming constraints. \textit{Hierarchical Memory} organises retained evidence in a \textit{Parallel Representation} that aligns temporal context with spatial observations. \textit{Agentic Retrieval} searches, inspects, and verifies candidate evidence before producing a grounded answer. On OVO-Bench, VAM achieves the highest RT+BT average (68.41) across all reported baselines, improving over end-to-end use of the same underlying MLLM (Gemini 3 Flash, 67.46). On the month-scale split of MM-Lifelong \textit{train@month} (105.6 hours over 51 days), VAM reaches 17.11\%, second only to ReMA with GPT-5 (17.62\%). These results suggest that long-horizon video understanding benefits from treating visual memory as an explicit, inspectable, and queryable substrate. Code is available at \url{https://github.com/yiliu-li/Visual-Agentic-Memory}.

%% file: sections/introduction.tex
\section{Introduction} \label{Sec:Introduction}

A central challenge in long video understanding is that high-bandwidth video capture is now practical, yet turning unbounded streams into a stable, queryable, and verifiable memory substrate remains difficult. Modern multimodal systems can process substantial visual context, but they still struggle to maintain a usable account of what happened, where it happened, and how scenes evolved over time \cite{grauman2022ego4dworld3000hours,yang2025egolifeegocentriclifeassistant,chen2026multimodallifelongunderstandingdataset,du2024towards}. A system cannot reason about the future or explain the past if earlier observations cannot be retained, localised, and revisited reliably \cite{ha2018worldmodels,zhu2024is,team2026openworldlib,hafner2023mastering,nvidia2025cosmos}. The practical question is how archives spanning hours, days, or months can remain searchable, auditable, and useful for downstream reasoning under real-world constraints \cite{lin2025streaming,11094230,song2023moviechat,10.5555/3737916.3738823,11094860}.

Existing paradigms address only part of this problem. Direct long-context models remain vulnerable to attention dilution as duration grows \cite{zou2024from,chen2024longvila}. Compact memory schemes improve tractability but often weaken recoverability once fine-grained evidence has been discarded. This creates a recoverability collapse in which compressed memory preserves semantic gist while discarding the visual evidence needed for later verification \cite{xie2026fluxmem,song2024moviechat+,Ma_2025_CVPR}. Recursive agentic approaches can also accumulate memory drift when intermediate reasoning cannot be checked against preserved observations \cite{chen2026multimodallifelongunderstandingdataset}. Online-tuned designs often compress aggressively or maintain only shallow recent memory \cite{liu2026thinkingstreamingvideo,wang2025streambridge}, whereas long-range-tuned designs often rely on offline preprocessing or memory structures that are costly to update continuously \cite{rege2026agentic,VideoAgent,wen2026eventmemagent}. These difficulties motivate a unified view of the problem as the orchestration of indexing, storage, and retrieval rather than longer context alone.

Strong pretrained multimodal models offer broad semantic priors but do not by themselves solve evidence retention, temporal disambiguation, provenance tracking, or long-horizon retrieval \cite{alayrac2022flamingovisuallanguagemodel,li2023blip2bootstrappinglanguageimagepretraining,bai2025qwen3vltechnicalreport,qwen3_5_2026}. We therefore formulate long video understanding around three coupled requirements. \textit{Online Indexing} retains visual evidence across temporal and spatial dimensions under streaming constraints \cite{xie2026fluxmem,zeng2025streamforest}. \textit{Hierarchical Memory} keeps timeline context aligned with recoverable raw observations rather than only compressed summaries \cite{song2023moviechat,rege2026agentic}. \textit{Agentic Retrieval} performs task-relevant search and visual inspection over that memory \cite{videoexplorer2026,pan2025timesearchradaptivetemporalsearch,fan2025videoagent,VideoAgent}. This framing also distinguishes retrieval-first agentic systems from recursive approaches such as ReMA, where iterative reasoning plays a larger role than explicit evidence retrieval \cite{chen2026multimodallifelongunderstandingdataset}.

To address these constraints, we propose \textit{Visual Agentic Memory} (VAM), a training-free framework that transforms raw video streams into searchable long-horizon memory while preserving the original frames as first-class evidence. In our MM-Lifelong evaluation, this design scales to month-long streams (51 days). VAM updates memory incrementally as frames arrive and avoids expensive global preprocessing such as optical flow \cite{xie2026fluxmem,wen2026eventmemagent}. It organises retained moments into events through a parallel temporal-spatial representation. Each event maintains a temporal representation for timeline reasoning and a spatial representation containing retained raw frames and embeddings for later inspection. The same MLLM family writes event summaries from the temporally bounded evidence, capturing the salient activity, entities, and temporal cues within each event. At query time, that model family also serves as the retrieval agent. It performs multi-turn search and visual inspection until it can answer with grounded evidence and cited intervals \cite{pan2025timesearchradaptivetemporalsearch,tstar}.

\begin{figure}[t]
\centering
\includegraphics[width=\textwidth]{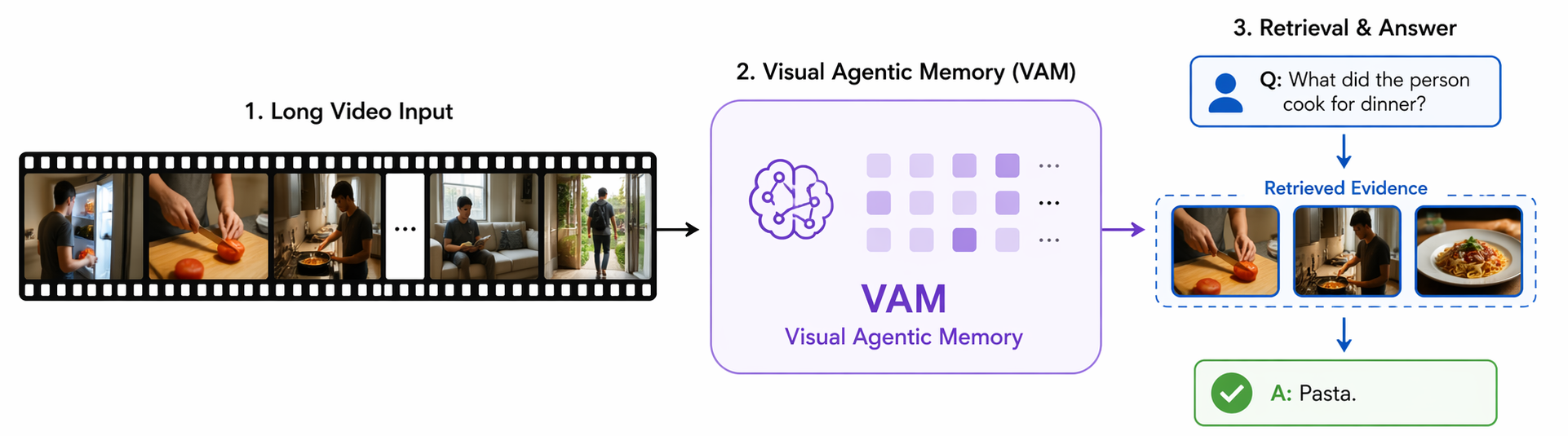}
\caption{VAM turns continuous video streams into searchable long-horizon memory. Unlike prior offline methods that operate on finite clips and suffer from memory explosion, VAM supports online indexing and agentic retrieval over effectively unbounded video through adaptive temporal-visual compression, hierarchical memory, and parallel temporal-spatial representations.}
\label{fig:teaser}
\end{figure}

Our contributions are as follows.
(1) We formulate online long-video understanding as an evidence-recoverable visual memory problem, shifting the focus from longer context windows to searchable and inspectable memory.
(2) We propose VAM, a training-free retrieval-first framework that incrementally builds memory from streaming video and decouples continuous indexing from query-time reasoning.
(3) We introduce a hierarchical parallel temporal-spatial representation that aligns event-level summaries with retained raw frames, supporting both efficient timeline search and direct visual verification.
(4) We demonstrate that VAM improves online and long-horizon video understanding, achieving the highest RT+BT average on OVO-Bench and competitive accuracy on MM-Lifelong \textit{train@month}, a month-scale benchmark (105.6 hours, 51 days), while storing only a small fraction of the raw stream.

%% file: sections/related_work.tex
\section{Related Work} \label{Sec:RelatedWork}

Online long video understanding must update memory continuously under streaming input and keep it retrievable over long horizons. As duration extends from minutes to months, evidence becomes sparse, repeated events become harder to disambiguate, and retrieval errors propagate more severely \cite{10.5555/3737916.3738823,11094860,song2024moviechat+}.

\subsection{Methods}
\label{SubSec:Methods}
\begin{figure}[t]
\centering
\includegraphics[width=\textwidth]{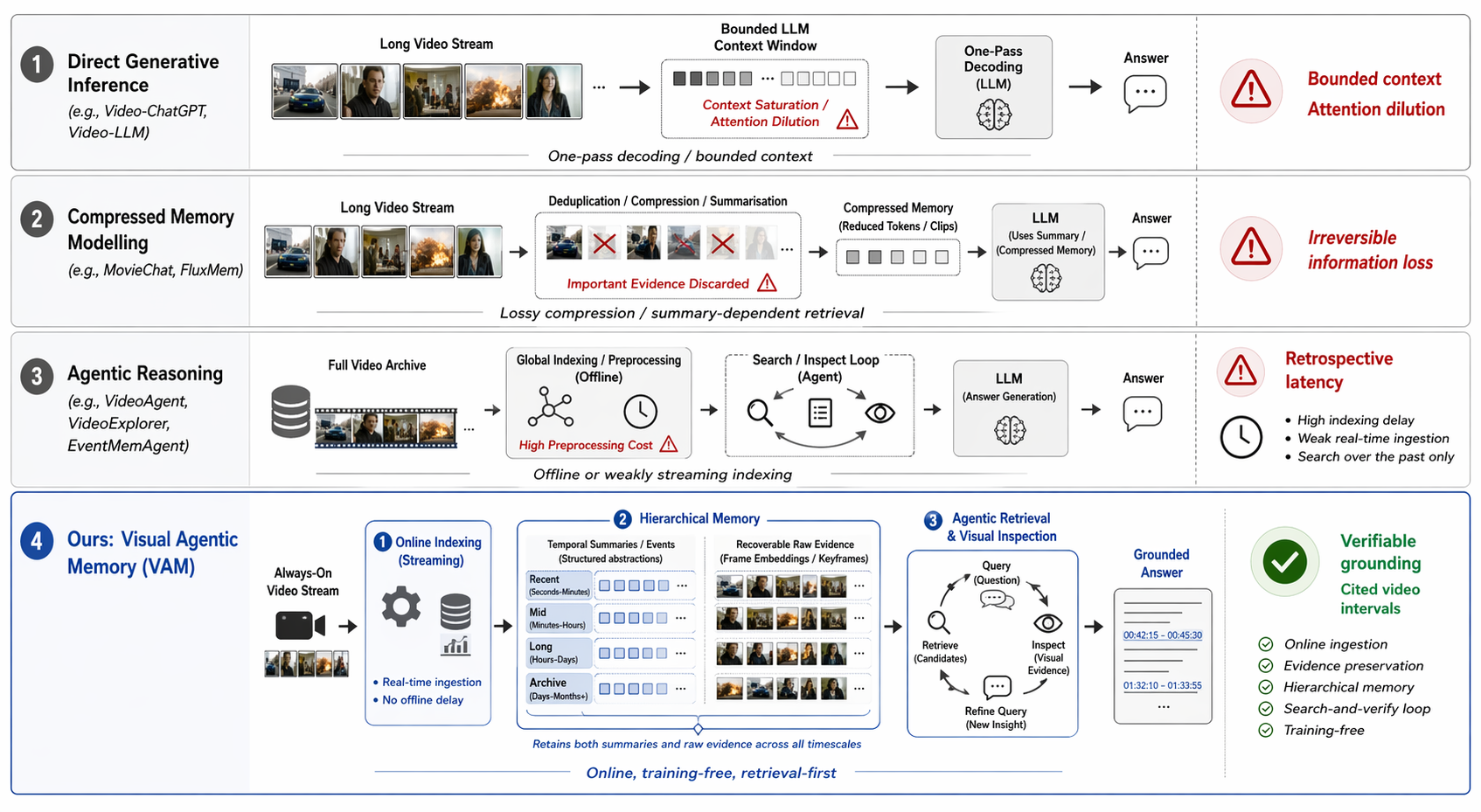}
\caption{Three macro-paradigms in online long video understanding: direct generative inference, compressed memory modelling, and agentic reasoning. VAM occupies a retrieval-first design point combining online indexing with explicit agentic retrieval over unbounded streams.}
\label{fig:paradigm_comparison}
\end{figure}

Building on increasingly capable video-native MLLMs \cite{maaz2024videochatgptdetailedvideounderstanding,wang2024internvideo2,gemini_video_understanding_2025,gemini3_2025,bai2025qwen3vltechnicalreport,qwen3_5_2026}, current systems fall into three macro-paradigms (Figure~\ref{fig:paradigm_comparison}).

\textbf{Direct generative inference.} Systems decode answers from in-context visual tokens \cite{lin-etal-2024-video,maaz2024videochatgptdetailedvideounderstanding,chen2024imageworth12tokens}. They are typically paired with fixed-rate sampling or bounded-context prompting, which saturates context as duration grows and remains vulnerable to attention dilution \cite{zou2024from}. Streaming refinements such as ViSpeak \cite{fu2025vispeak} and StreamAgent \cite{yang2025streamagentanticipatoryagentsstreaming} strengthen real-time interaction or task anticipation, but stop short of a persistent evidence-preserving architecture.

\textbf{Compressed memory modelling.} These methods retain compact memory rather than all recoverable evidence. Representative examples include MovieChat(+) with sparse retention and summarisation \cite{song2023moviechat,song2024moviechat+}, StreamForest with a progressively compressed hierarchy \cite{zeng2025streamforest}, Infinity-Video with continuous-time consolidation \cite{santos2025inftyvideotrainingfreeapproachlong}, FluxMem with Otsu-derived adaptive thresholds for spatial--temporal redundancy removal \cite{xie2026fluxmem}, and StreamMind with online memory compression \cite{ding2025streammind,liu2026thinkingstreamingvideo}. They trade evidence fidelity for bounded storage. Once fine-grained details are discarded, later reasoning cannot revisit the original observations, weakening citation fidelity and retrospective inspection.

\textbf{Agentic reasoning.} These systems interleave planning, search, inspection, and recursive refinement through iterative tool use \cite{rege2026agentic,VideoAgent,videoexplorer2026,wen2026eventmemagent,chen2026multimodallifelongunderstandingdataset,arnab2025temporalchainthoughtlongvideo,zhang2025videoarm,yeo2026worldmm,chen2026telemem,wang2024lifelongmemory}. They differ markedly in their primary substrate. Examples include graph-structured semantic memory (EGAgent \cite{rege2026agentic}), multi-agent collaboration (LVAgent \cite{chen2025lvagentlongvideounderstanding}), trained ``thinking with video'' policies (VideoExplorer \cite{videoexplorer2026}), hierarchical event-centric memory (EventMemAgent \cite{wen2026eventmemagent}), adaptive reasoning over dynamically constructed hierarchical memory (VideoARM \cite{zhang2025videoarm}), multimodal memory combining episodic, semantic, and visual representations (WorldMM \cite{yeo2026worldmm}), long-term multimodal memory for agentic systems (TeleMem \cite{chen2026telemem}), LLM-based reasoning over compressed textual descriptions (LifelongMemory \cite{wang2024lifelongmemory}), and recursive belief updating (ReMA, Temporal Chain of Thought \cite{chen2026multimodallifelongunderstandingdataset,arnab2025temporalchainthoughtlongvideo}). Tool use does not imply evidence preservation. Most agents do not organise raw RGB observations as a first-class queryable substrate, exposing them to global localisation collapse as the searchable pool grows \cite{chen2026multimodallifelongunderstandingdataset}. Recursive variants are also sensitive to early hypothesis errors that are inherited rather than corrected through direct lookup. A retrieval-first design point, as adopted in this work, explicitly retains raw frames alongside event summaries, enabling direct visual verification rather than relying solely on compressed or textual representations.

\subsection{Benchmarks}
\label{SubSec:Benchmarks}
Long video benchmarks fall into \textit{online} benchmarks (timestamp-conditioned, sequential input) and \textit{long-horizon} benchmarks (retrieval and reasoning over extended context). Beyond short-clip predecessors like MovieQA and TVQA \cite{tapaswi2016movieqaunderstandingstoriesmovies,lei-etal-2018-tvqa}, recent coverage spans StreamingBench, LongVideoBench, MLVU, EgoLifeQA, Video-MME, OVO-Bench, and MM-Lifelong \cite{lin2025streaming,10.5555/3737916.3738823,11094860,yang2025egolifeegocentriclifeassistant,fu2025video,11094230,chen2026multimodallifelongunderstandingdataset}. We use \textit{OVO-Bench} \cite{11094230} for strict online evaluation and \textit{MM-Lifelong train@month} \cite{chen2026multimodallifelongunderstandingdataset} for the long-horizon setting. The latter spans 105.6 hours over 51 days with a human--model gap exceeding 70 percentage points, explicitly exposing the working-memory bottleneck of end-to-end MLLMs and global localisation collapse in agentic baselines.

%% file: sections/methodology.tex
\section{Methodology} \label{Chap:Methodology}

The VAM framework transforms a continuous video stream into a searchable memory trace for retrieval, inspection, and citation. The objective is to retain recoverable visual evidence while keeping long-horizon search tractable. To do this, continuous memory construction is decoupled from query-time reasoning, so indexing proceeds as the stream evolves.

\subsection{System Architecture}
\label{Sec:SystemArchitecture}
\begin{figure}[!b]
\centering
\includegraphics[width=\textwidth]{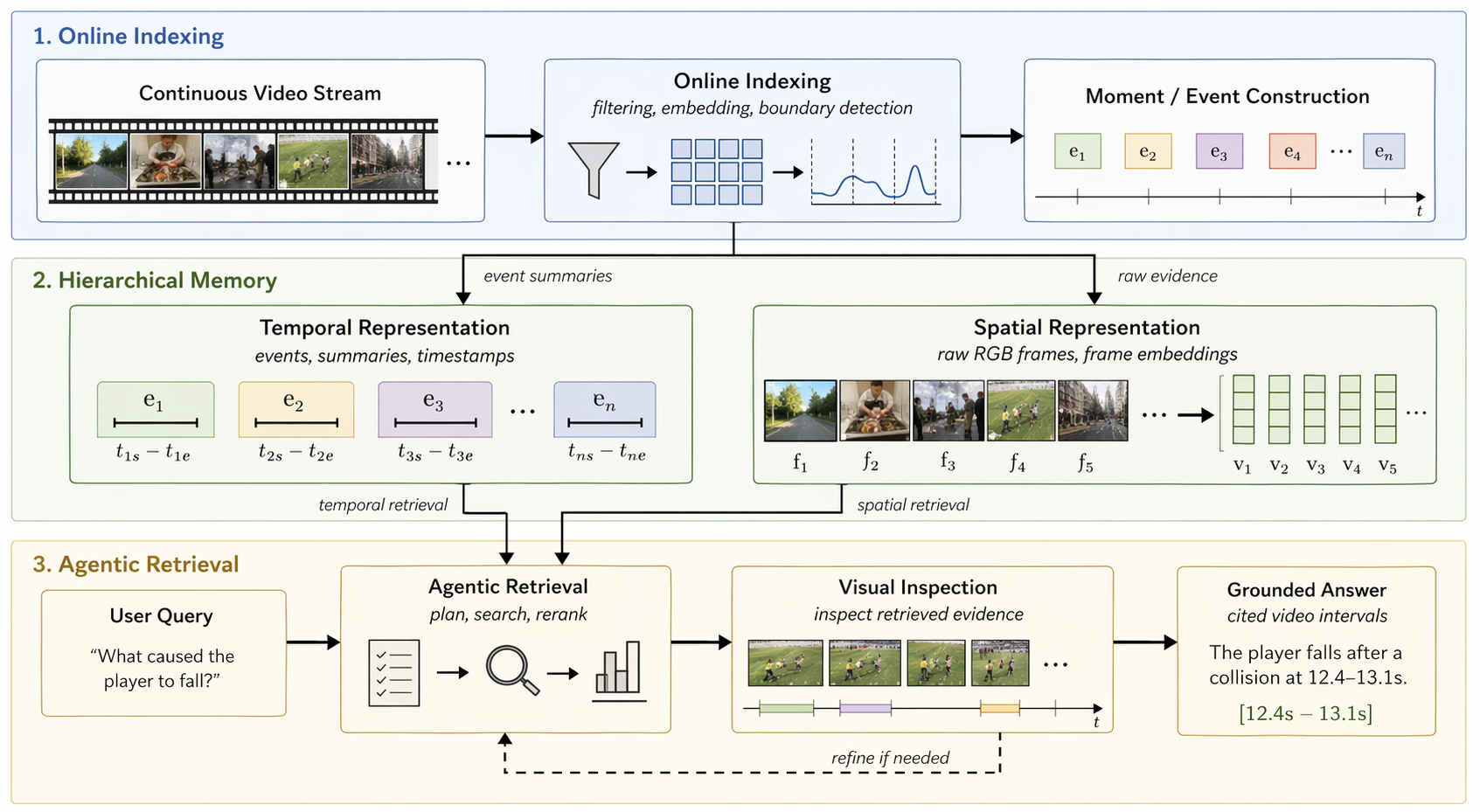}
\caption{System architecture. Three coupled layers (Online Indexing, Hierarchical Memory, and Agentic Retrieval) feed a parallel temporal/spatial memory. At query time the agent retrieves from both representations and inspects raw frames before answering. The dashed link shows feedback driving additional retrieval.}
\label{fig:architecture}
\end{figure}

The framework consists of three coupled components: online indexing (Section~\ref{Sec:OnlineIndexing}), a hierarchical memory substrate (Section~\ref{Sec:HierarchicalMemory}), and query-time agentic retrieval (Section~\ref{Sec:AgenticRetrieval}). These components are joined by asynchronous orchestration (Section~\ref{Sec:AsynchronousOrchestration}). Online indexing filters incoming frames and retains visually significant transitions as \textit{moments}, which are frame-level units storing raw observations and embeddings. Contiguous moments are grouped into \textit{events} carrying event summaries and temporal metadata. Event summaries support coarse retrieval, while spatial evidence remains available for direct inspection. At query time, an MLLM acts as the retrieval agent. It iteratively consults both pathways and synthesises answers anchored to recoverable visual intervals. The same model family writes event summaries during indexing, so summarisation and inspection are two roles of a shared MLLM interface.

\subsection{Online Indexing}
\label{Sec:OnlineIndexing}
Online indexing transforms the dense raw stream into a sparser yet evidence-preserving memory trace through three consecutive functions. The first filters observations that are unlikely to support later reasoning. The second encodes retained frames for retrieval. The third commits memory boundaries that preserve transitions while compacting quasi-static intervals. Given $\mathcal{F} = \{F_1, \dots, F_N\}$ with relative and absolute timestamps $t_i, T_i$, the resulting frame-level entries are \textit{moments}, each storing a retained frame with embedding and timestamps. Contiguous moments later define \textit{events}.

\subsubsection{Stage 1: Frame Filtering}
The filtering stage removes observations that are visually unreliable or insufficiently novel before any high-cost neural processing.

\textbf{Blur Detection (Laplacian Variance).} Sharpness is the variance of the Laplacian operator:
\begin{equation}
    \text{Sharpness}(I_t) = \text{Var}(\nabla^2 I_{t,\mathrm{gray}})
\end{equation}
A blurry image yields lower variance. If $\text{Sharpness}(I_t) < \tau_{blur}$, the frame is discarded.

\textbf{Redundancy Filtering.} Many streams remain dominated by visually near-identical frames even when image quality is acceptable. The system performs pre-embedding redundancy suppression based on visual similarity, placed before dense embedding so repeated content is discarded early. Only frames passing both gates proceed (Figure~\ref{fig:prefiltering_example}, defaults in Appendix~\ref{app:implementation}).

\begin{figure}[htbp]
\centering
\includegraphics[width=1.0\textwidth]{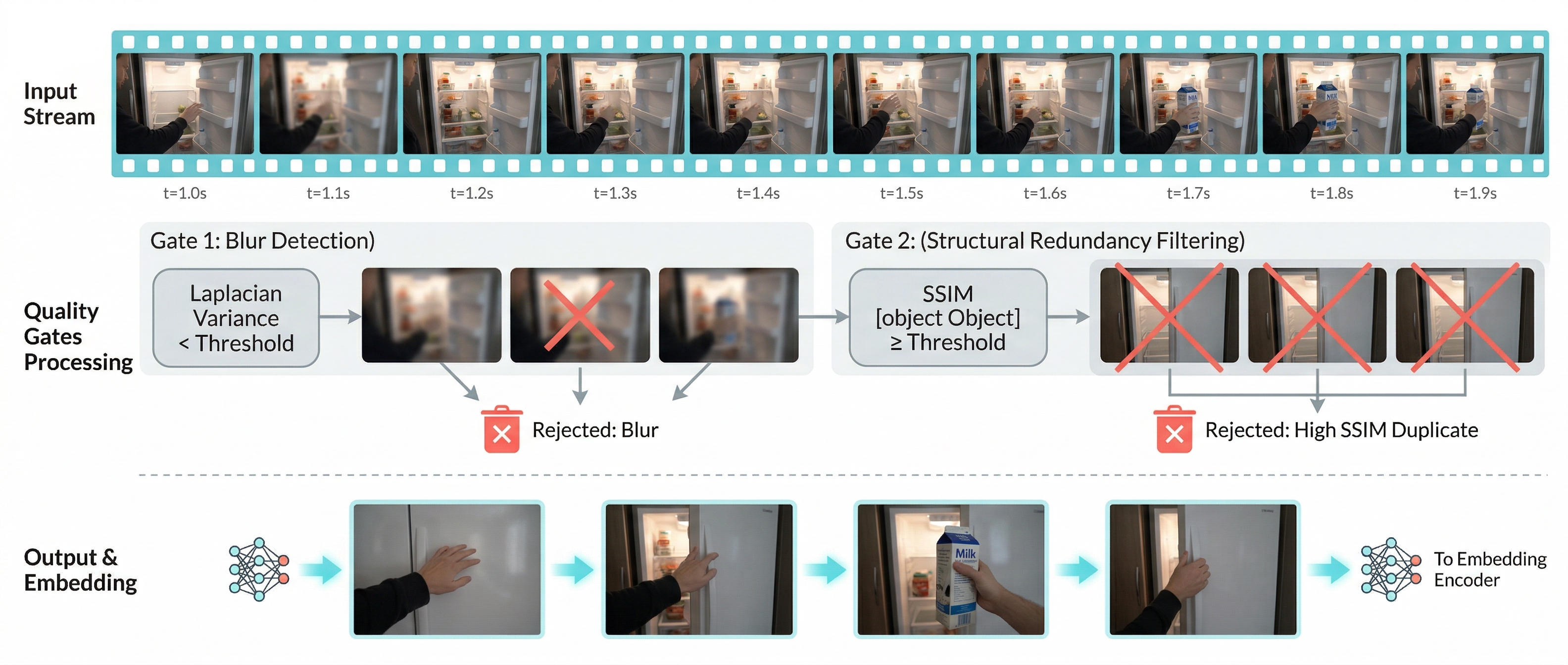}
\caption{Frame filtering. Top: raw stream. Middle: frames rejected for blur ($< \tau_{blur}$) or redundancy. Bottom: retained keyframes proceeding to embedding.}
\label{fig:prefiltering_example}
\end{figure}

\subsubsection{Stage 2: Embedding Extraction}
For each retained frame, the system extracts a dense embedding $\mathbf{e}_t \in \mathbb{R}^{d}$ using a pre-trained multimodal embedding model \cite{radford2021learningtransferablevisualmodels,Bain21}. The specific instantiation is an experimental choice. This embedding model serves as the retrieval-facing representation layer. It supports boundary-sensitive change detection, frame-level retrieval, and query--evidence alignment. The MLLM is reserved for event summarisation, visual inspection, and answer synthesis.

\subsubsection{Stage 3: Memory Update and Boundary-Preserving Moment Commit}
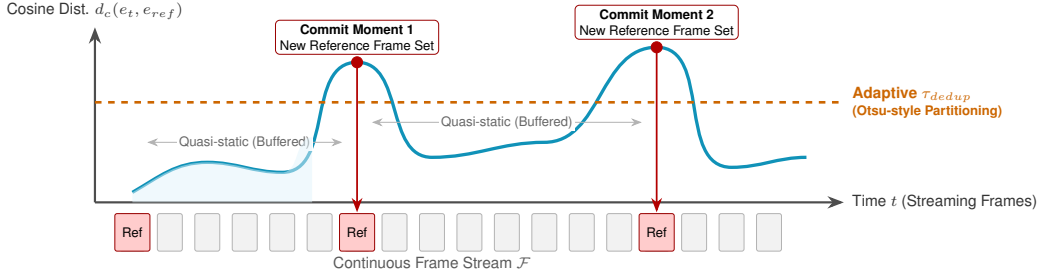
\begin{figure}[htbp]
\centering
\resizebox{1.0\textwidth}{!}{
\begin{tikzpicture}[
    x=1.2cm, y=0.8cm,
    font=\sffamily,
    >=Stealth,
    ax/.style={->, line width=1.0pt, draw=black!70},
    curve/.style={line width=1.5pt, draw=cyan!70!black, smooth},
    thresh/.style={line width=1.2pt, draw=orange!80!black, dashed},
    lbl/.style={font=\sffamily\scriptsize, text=black!80},
    callout/.style={rectangle, fill=white, draw=gray!40, rounded corners=2pt, font=\sffamily\tiny, inner sep=2pt, align=center},
    framebox/.style={rectangle, draw=gray!60, fill=gray!10, minimum width=0.4cm, minimum height=0.6cm, rounded corners=1pt, font=\sffamily\tiny}
]
\draw[ax] (0, 0) -- (10, 0) node[right, lbl] {Time $t$ (Streaming Frames)};
\draw[ax] (0, 0) -- (0, 3.5) node[above, lbl] {Cosine Dist. $d_c(e_t, e_{ref})$};
\foreach \x in {0.5, 1.0, 1.5, 2.0, 2.5, 3.0, 3.5, 4.0, 4.5, 5.0, 5.5, 6.0, 6.5, 7.0, 7.5, 8.0, 8.5, 9.0} {
    \node[framebox] at (\x, -0.6) {};
}
\node[framebox, fill=red!20, draw=red!60!black] at (0.5, -0.6) {Ref};
\node[framebox, fill=red!20, draw=red!60!black] at (3.5, -0.6) {Ref};
\node[framebox, fill=red!20, draw=red!60!black] at (7.5, -0.6) {Ref};
\node[font=\sffamily\scriptsize, text=gray!70!black] at (4.5, -1.2) {Continuous Frame Stream $\mathcal{F}$};
\draw[curve] (0.5, 0.2)
    to[out=30, in=180] (1.5, 0.8)
    to[out=0, in=180] (2.5, 0.6)
    to[out=0, in=180] (3.5, 2.8)
    to[out=0, in=180] (4.5, 0.9)
    to[out=0, in=180] (6.0, 1.2)
    to[out=0, in=180] (7.5, 3.1)
    to[out=0, in=180] (8.5, 0.7)
    to[out=0, in=180] (9.5, 0.9);
\draw[thresh] (0, 2.0) -- (10, 2.0) node[right, font=\sffamily\scriptsize\bfseries, text=orange!80!black, align=left] {Adaptive $\tau_{dedup}$\\\tiny (Otsu-style Partitioning)};
\filldraw[red!70!black] (3.5, 2.8) circle (2.5pt);
\draw[->, draw=red!70!black, line width=0.8pt] (3.5, 2.8) -- (3.5, -0.2);
\node[callout, draw=red!60!black] at (3.5, 3.3) {\textbf{Commit Moment 1}\\New Reference Frame Set};
\filldraw[red!70!black] (7.5, 3.1) circle (2.5pt);
\draw[->, draw=red!70!black, line width=0.8pt] (7.5, 3.1) -- (7.5, -0.2);
\node[callout, draw=red!60!black] at (7.5, 3.6) {\textbf{Commit Moment 2}\\New Reference Frame Set};
\draw[<->, draw=gray!60] (0.7, 1.2) -- (3.3, 1.2) node[midway, fill=white, font=\sffamily\tiny, text=gray!80!black, inner sep=1pt] {Quasi-static (Buffered)};
\draw[<->, draw=gray!60] (3.7, 1.6) -- (7.3, 1.6) node[midway, fill=white, font=\sffamily\tiny, text=gray!80!black, inner sep=1pt] {Quasi-static (Buffered)};
\fill[cyan!10, opacity=0.5] (0.5,0) -- (0.5,0.2)
    to[out=30, in=180] (1.5, 0.8)
    to[out=0, in=180] (2.5, 0.6)
    to[out=0, in=180] (2.9, 1.3) -- (2.9, 0) -- cycle;
\end{tikzpicture}
}
\caption{Adaptive deduplication. When $d_c$ exceeds the adaptive $\tau_{dedup}$, buffered moments are committed and the reference is reset. The same distance spikes drive event formation.}
\label{fig:adaptive_dedup}
\end{figure}

The system decides whether a new frame provides genuinely new information by comparing each incoming embedding to the current reference:
\begin{equation}
    d_c(\mathbf{e}_t, \mathbf{e}_{ref}) = 1 - \frac{\mathbf{e}_t \cdot \mathbf{e}_{ref}}{\|\mathbf{e}_t\| \|\mathbf{e}_{ref}\|}
\end{equation}
A fixed threshold is brittle because embedding variation depends on scene dynamics, camera motion, and environmental stability. VAM therefore maintains a sliding distance history $\mathcal{H}_d$ and derives an adaptive deduplication gate from recent statistics. With sufficient samples, Otsu partitioning \cite{4310076} is applied over $\mathcal{H}_d$. Otherwise, a configured similarity-derived default is used. This is methodologically close to FluxMem \cite{xie2026fluxmem}, which applies Otsu thresholds to token reduction in temporal adjacency and spatial domain consolidation rather than to evidence-preserving indexing:
\begin{equation}
    \tau_{dedup} \approx \text{AdaptiveThreshold}(\mathcal{H}_d).
\end{equation}
Frames with $d_c \le \tau_{dedup}$ are buffered as redundant context. When $d_c > \tau_{dedup}$, the buffered endpoint is committed as a moment and the new frame becomes the reference \cite{rege2026agentic,videoexplorer2026}. Figure~\ref{fig:adaptive_dedup} illustrates this update rule. The same Otsu-based change analysis defines event boundaries: distances between consecutive moments are scanned for local peaks above a relaxed threshold, with a duration cap preventing unbounded events.

At runtime each frame undergoes (1) encoding to $\mathbf{e}_t$, (2) computing $d_c$ against the reference, (3) comparing with $\tau_{dedup}$, and (4) buffering or committing. The process is strictly online, with no future-frame access \cite{rege2026agentic,videoexplorer2026}.

\subsection{Hierarchical Memory}
\label{Sec:HierarchicalMemory}
Embeddings alone are insufficient for episodic retrieval over long timelines. Precise frame matching and abstract event recall fail under different conditions \cite{song2023moviechat,rege2026agentic,videoexplorer2026,wen2026eventmemagent}. Hierarchical memory organises moments into events so spatial and temporal representations stay synchronised through shared anchors.

\textbf{Age-aware storage.} Retained items occupy \textit{recent}, \textit{mid}, and \textit{long} tiers by age $\Delta t$. Compression strength and minimum spacing increase with age: recent windows retain dense full-frame payloads, older segments are consolidated into compact event-level cues that remain retrievable \cite{rege2026agentic,videoexplorer2026}.

\textbf{Event formation.} Boundaries come from the same Otsu-derived statistics applied to the moment-distance sequence, with a maximum event duration. Letting $m_j$ denote the $j$-th moment, consecutive moments between two boundaries form an event:
\begin{equation}
    E_i = \{m_{s_i}, \dots, m_{e_i}\}
\end{equation}
The MLLM generates an event summary from the same temporally bounded evidence (a clip cropped over the event interval when available, otherwise representative retained frames with metadata):
\begin{equation}
    C_i = \text{MLLM}(E_i)
\end{equation}
Each event maintains (i) retained moments, (ii) embeddings $\mathbf{e}_t$, (iii) summary $C_i$, and (iv) metadata $(t, T)$. The spatial representation comprises moments and embeddings. The temporal representation consists of event summaries and time-aware retrieval cues. The two pathways share metadata, so retrieval can move from event-level temporal hypotheses to frame-level spatial evidence without alignment loss. Broader \textit{summary documents} are materialised at query time by aggregating event summaries rather than replacing them.

\subsection{Agentic Retrieval}
\label{Sec:AgenticRetrieval}
Agentic retrieval is a bounded multi-turn loop. At each turn $k$, the MLLM receives the query $Q$ and history $\mathcal{C}_{k-1}$, and selects an action:
\[
 a_k \in \{\texttt{search}, \texttt{inspect}, \texttt{summarize}, \texttt{answer}\}.
\]
The loop follows the observe--hypothesise--inspect paradigm \cite{rege2026agentic,videoexplorer2026}. This paradigm constrains inference to interpretable primitives, which preserves auditability and stabilises long-horizon search.

\paragraph{Intent parsing.} Each turn converts the query into an operational specification. For complex requests, the MLLM decomposes user intent into executable sub-queries with explicit temporal constraints, target entities, operator selection, and source prioritisation.

\paragraph{Hybrid temporal-spatial retrieval.} \texttt{search} consults two pathways within a hybrid retrieval step. Let $\mathbf{q}_k$ denote the turn-$k$ retrieval embedding and $D_{TEMP} = \{d_i\}$ the set of event summaries (or materialised summary documents) with embeddings $\mathbf{v}_{d_i}$. \emph{Temporal retrieval} scores summaries via cosine similarity, $S_{TEMP}(d_i) = \cos(\mathbf{q}_k, \mathbf{v}_{d_i})$, prioritising situational context. Summary documents aggregate overlapping evidence into a single retrieval-oriented description because retrieval over local event summaries alone does not guarantee cross-event consolidation. \emph{Spatial retrieval} compares the query against frame embeddings, $S_{VIS}(F_t) = \cos(\mathbf{q}_k, \mathbf{e}_t)$, pinpointing object-level details. The agent can also use \texttt{visual\_ref} for appearance-based retrieval. In addition, \texttt{inspect\_k} enables look-ahead by automatically passing top candidates to inspection.

\paragraph{Grounded inspection.} Retrieved candidates remain provisional. The MLLM invokes \texttt{inspect} (optionally as \texttt{joint\_inspection} when comparing temporally related candidates) to evaluate evidence directly:
\begin{equation}
    O_k = \text{Inspector}(\mathcal{X}^{vis}_k, \mathcal{X}^{temp}_k, Q)
\end{equation}
where $\mathcal{X}^{vis}_k, \mathcal{X}^{temp}_k$ are the candidates selected at turn $k$. Inspection resolves ambiguities, adjudicates conflicts between retrieval signals, and is the primary safeguard against unsupported answers. \texttt{summarize} synthesises localised event summaries and frames into a reusable summary document persisted in memory.

\paragraph{Answer synthesis and termination.} Termination is an explicit planner action. \texttt{answer} is emitted once the context contains sufficient evidence, the turn budget grows small, or successive turns cease to improve results. The answer must be associated with a \texttt{best\_ref}, typically a \texttt{(turn\_idx, result\_idx)} pointer to the most probative candidate, resolved against the stored turn context so the response remains coupled to a concrete evidence instance. When \texttt{best\_ref} is absent or invalid, the controller falls back to the highest-scoring retrieved candidate. If progress stagnates or the maximum turn budget $K$ is reached, termination is forced. The output preserves the answer text, evidence sufficiency, principal supporting candidate, the full multi-turn trace, and total search duration.

\subsection{Asynchronous Orchestration}
\label{Sec:AsynchronousOrchestration}
Online indexing and agentic retrieval are decoupled but coordinated processes over the shared memory. Indexing runs in the background, filtering frames and writing moments and events. Agentic retrieval is instantiated on demand and interacts with memory only through bounded retrieval operators \cite{lin2025streaming}. Explicit resource bounds (turn budgets, candidate set sizes, and inspection batch limits) keep retrieval cost predictable while cited evidence stays grounded in indexed memory entries.

A practical consequence is that long-horizon retrieval often requires passing multiple candidate intervals or frame sets to the MLLM for direct inspection. This makes evaluation substantially more expensive than retrieval-only or one-pass decoding settings. As in many recent long-video studies, the empirical scope is further constrained by inference budget and serving access for repeated multi-image MLLM calls. The study therefore adopts a targeted rather than exhaustive evaluation design, focusing on representative benchmark subsets that are most informative for online responsiveness, retrospective evidence recovery, and grounded answer generation.

%% file: sections/evaluation.tex
\section{Evaluation} \label{Sec:Evaluation}

VAM is evaluated along two directions: online performance under strict streaming constraints (\textit{OVO-Bench}) and long-horizon performance over extended temporal spans (the month-scale portion of \textit{MM-Lifelong}). Across all results, VAM uses Gemini Embedding 2 for memory indexing and retrieval, Gemini 3 Flash as the query-time MLLM, and an initial 0.5 fps sampling rate. Retained memory is then formed adaptively rather than stored at a constant fps.

\subsection{OVO-Bench Results}
\label{SubSec:PerformanceAxes}
\textit{OVO-Bench} \cite{11094230} contains 644 videos and 3100 queries, with sources ranging from minutes to roughly half an hour. It tests timestamp-conditioned operation in which each response is grounded only in observations available up to the query time. It decomposes online performance into \textit{Real-Time Visual Perception} (RT), \textit{Backward Tracing} (BT), and \textit{Forward Active Responding} (Fwd). RT and BT are the most informative components for VAM, since they test whether online indexing preserves a usable memory substrate and whether agentic retrieval can localise earlier evidence within it.

\begin{table}[tbp]
\centering
\resizebox{\textwidth}{!}{
\begin{tabular}{lccccccc|cccc|cccc|cc}
\toprule
\multirow{2}{*}{Model} & \multicolumn{7}{c}{Real-Time Visual Perception} & \multicolumn{4}{c}{Backward Tracing} & \multicolumn{4}{c}{Forward Active Responding} & \multirow{2}{*}{\shortstack{RT+BT\\Avg.}} & \multirow{2}{*}{\shortstack{Overall}} \\
\cmidrule(lr){2-8}\cmidrule(lr){9-12}\cmidrule(lr){13-16}
 & OCR & ACR & ATR & STU & FPD & OJR & \textbf{Avg.} & EPM & ASI & HLD & \textbf{Avg.} & REC & SSR & CRR & \textbf{Avg.} & & \\
\midrule
Human Agents & 93.96 & 92.57 & 94.83 & 92.70 & 91.09 & 94.02 & 93.20 & 92.59 & 93.02 & 91.37 & 92.33 & 95.48 & 89.67 & 93.56 & 92.90 & 92.77 & 92.81 \\
\midrule
\multicolumn{18}{c}{\textbf{Proprietary MLLMs}} \\
\midrule
Gemini 3 Flash \cite{gemini3_2025} & \textbf{86.58} & 68.81 & \underline{79.31} & \underline{58.99} & 70.30 & 70.65 & \underline{72.44} & 58.92 & \textbf{76.35} & 52.15 & \textbf{62.47} & 35.53 & \textbf{74.24} & 61.67 & \textbf{57.15} & \underline{67.46} & \textbf{64.02} \\
GPT-4o \cite{openai2024gpt4o} & 69.80 & 64.22 & 71.55 & 51.12 & 70.30 & 59.78 & 64.46 & 57.91 & \underline{75.68} & 48.66 & 60.75 & 27.58 & \underline{73.21} & 59.40 & 53.40 & 62.61 & 59.54 \\
\midrule
\multicolumn{18}{c}{\textbf{Open-Source Offline MLLMs}} \\
\midrule
Qwen2-VL-72B \cite{wang2024qwen2vl} & 65.77 & 60.55 & 69.83 & 51.69 & 69.31 & 54.35 & 61.92 & 52.53 & 60.81 & \underline{57.53} & 56.96 & \underline{38.83} & 64.07 & 45.00 & 49.30 & 59.44 & 56.06 \\
LLaVA-Video-7B \cite{zhang2024llavavideo} & 69.13 & 58.72 & 68.83 & 49.44 & 74.26 & 59.78 & 63.36 & 56.23 & 57.43 & 7.53 & 40.40 & 34.10 & 69.95 & 60.42 & 54.82 & 51.88 & 52.86 \\
LLaVA-OneVision-7B \cite{li2024llavaonevision} & 66.44 & 57.80 & 73.28 & 53.37 & 71.29 & 61.96 & 64.02 & 54.21 & 55.41 & 21.51 & 43.71 & 25.64 & 67.09 & 58.75 & 50.49 & 53.86 & 52.74 \\
Qwen2-VL-7B \cite{wang2024qwen2vl} & 60.40 & 50.46 & 56.03 & 47.19 & 66.34 & 55.43 & 55.97 & 47.81 & 35.48 & 56.08 & 46.46 & 31.66 & 65.82 & 48.75 & 48.74 & 51.22 & 50.39 \\
InternVL-V2-8B \cite{chen2024internvl2} & 67.11 & 60.55 & 63.79 & 46.07 & 68.32 & 56.52 & 60.39 & 48.15 & 57.43 & 24.73 & 43.44 & 26.50 & 59.14 & 54.14 & 46.59 & 51.92 & 50.14 \\
LongVU-7B \cite{shen2024longvu} & 53.69 & 53.21 & 62.93 & 47.75 & 68.32 & 59.78 & 57.61 & 40.74 & 59.46 & 4.84 & 35.01 & 12.18 & 69.48 & 60.83 & 47.50 & 46.31 & 46.71 \\
Qwen3-VL-8B \cite{bai2025qwen3vltechnicalreport} & 75.17 & 58.72 & 72.41 & 57.30 & 70.30 & 59.24 & 65.52 & 56.57 & 69.59 & 12.37 & 46.18 & \textbf{47.13} & 67.57 & 52.50 & 55.73 & 55.85 & 55.81 \\
\midrule
\multicolumn{18}{c}{\textbf{Open-Source Online MLLMs/Agents}} \\
\midrule
Flash-VStream-7B \cite{zhang2025flashvstream} & 24.16 & 29.36 & 28.45 & 33.71 & 25.74 & 28.80 & 28.37 & 39.06 & 37.16 & 5.91 & 27.38 & 8.02 & 67.25 & 60.00 & 45.09 & 27.88 & 33.61 \\
VideoLLM-online-8B \cite{videollmonline2024} & 8.05 & 23.85 & 12.07 & 14.04 & 45.54 & 21.20 & 20.79 & 22.22 & 18.80 & 12.18 & 17.73 & - & - & - & - & 19.26 & - \\
Dispider \cite{qian2025dispider} & 57.72 & 49.54 & 62.07 & 44.94 & 61.39 & 51.63 & 54.55 & 48.48 & 55.41 & 4.30 & 36.06 & 18.05 & 37.36 & 48.75 & 34.72 & 45.31 & 41.78 \\
Streamo-7B (2fps*) \cite{xia2025streaming} & 77.18 & 66.06 & 76.72 & 45.51 & 66.34 & \underline{72.83} & 67.44 & 55.56 & 58.11 & 33.87 & 49.18 & 30.84 & 57.55 & \textbf{82.50} & \underline{56.96} & 58.31 & 57.86 \\
ViSpeak-7B \cite{fu2025vispeak} & 75.17 & 58.72 & 71.55 & 51.12 & 74.26 & 66.85 & 66.28 & \underline{59.93} & 48.65 & \textbf{63.98} & 57.52 & 33.81 & 68.52 & 60.42 & 54.25 & 61.90 & 59.35 \\
FluxMem \cite{xie2026fluxmem} & 81.20 & 59.60 & 70.70 & 53.40 & \textbf{75.20} & 63.00 & 67.18 & - & - & - & - & - & - & - & - & - & - \\
ThinkStream-3B \cite{liu2026thinkingstreamingvideo} & 85.23 & 64.22 & 69.82 & 49.43 & 69.31 & 64.13 & 67.02 & 53.87 & 59.46 & 43.55 & 52.29 & - & - & - & - & 59.66 & - \\
StreamForest-7B \cite{zeng2025streamforest} & 68.46 & 53.21 & 71.55 & 47.75 & 65.35 & 60.87 & 61.20 & 58.92 & 64.86 & 32.26 & 52.01 & 32.81 & 70.59 & 57.08 & 53.49 & 56.61 & 55.57 \\
StreamAgent \cite{yang2025streamagentanticipatoryagentsstreaming} & 71.20 & 53.20 & 63.60 & 53.90 & 67.30 & 58.70 & 61.32 & 54.80 & 58.10 & 25.80 & 46.23 & 35.90 & 48.40 & 52.00 & 45.43 & 53.78 & 50.99 \\
EventMemAgent \cite{wen2026eventmemagent} & 75.84 & \underline{69.72} & 73.28 & 55.62 & 67.33 & 67.93 & 68.29 & 59.60 & 70.95 & 43.55 & 58.03 & 33.67 & 72.02 & 62.08 & 55.92 & 63.16 & \underline{60.75} \\
\midrule
\multicolumn{18}{c}{\textbf{Ours}} \\
\midrule
VAM (Ours) & \underline{85.23} & \textbf{76.15} & \textbf{81.90} & \textbf{82.19} & \underline{74.26} & \textbf{73.91} & \textbf{78.94} & \textbf{62.29} & 66.22 & 45.16 & \underline{57.89} & 22.78 & 58.82 & \underline{65.42} & 49.01 & \textbf{68.41} & 61.95 \\
\bottomrule
\end{tabular}
}
\caption{OVO-Bench \cite{11094230} results across all three modes. The Gemini 3 Flash baseline is reproduced under the OVO-Bench protocol. ``-'' marks scores not reported by the source. Best/second-best (excluding human) are \textbf{bold}/\underline{underlined}.}
\label{tab:ovobench}
\end{table}

As shown in Table~\ref{tab:ovobench}, VAM achieves the highest RT+BT average (68.41) on OVO-Bench across all reported baselines, leading the Real-Time average (78.94) and remaining competitive on Backward Tracing (57.89). It is particularly strong on Spatial Understanding (82.19) and Episodic Memory (62.29). The end-to-end Gemini 3 Flash baseline reaches 67.46 using the same underlying MLLM but without the memory architecture. The gap (+0.95) therefore reflects the contribution of online indexing, hierarchical memory, and agentic retrieval rather than model capacity. The gain is consistent across RT subcategories (ACR +7.34, ATR +2.59, STU +23.20, OJR +3.26), suggesting a systematic architectural benefit rather than a single-task artefact.

Forward Active Responding interprets differently from the other modes. It tests calibration under partial evidence, namely whether the system can withhold commitment until decisive future evidence appears. VAM performs less strongly here because its controller is explicitly evidence-seeking and citation-oriented. Once retrieval surfaces observations that appear sufficient, the agent tends to commit even when the stream has not yet revealed the decisive later cue. The same evidence-first design that aids retrospective grounding can therefore become over-eager in forward tasks, consistent with the gap between VAM's main-table lead and its lower Forward Active average.

\subsection{MM-Lifelong Results}
\label{SubSec:LongHorizon}
The month-scale \textit{Live Stream} portion of \textit{MM-Lifelong} \cite{chen2026multimodallifelongunderstandingdataset} is among the most demanding long-video benchmarks available. It spans 105.6 hours of continuous footage over 51 days. Relevant evidence is often separated by very large temporal gaps and surrounded by substantial distractor content, and the human--model accuracy gap exceeds 70 percentage points (82.5\% vs.\ $\leq$17.62\%). We report \textit{train@month} (266 questions). Following the benchmark protocol, end-to-end MLLMs are evaluated by uniform sparse sampling up to their maximum context capacity rather than dense ingestion of the full lifelong stream. Agentic baselines run under official defaults, with the model-alignment adjustment to ReMA reported by the benchmark authors for fairness. \textit{Ref@300} denotes quantised temporal grounding at a 300-second resolution. The timeline is discretised into 5-minute bins, and the predicted clue interval is scored against the annotated interval at that granularity.

VAM attains 17.11\% answer accuracy (Table~\ref{tab:mmlifelong_trainmonth}), the second-highest reported result. It is 0.51 percentage points below \textit{ReMA with GPT-5} (17.62\%) and substantially above all listed end-to-end MLLM baselines. Its Ref@300 (3.65) is more modest. Because VAM retrieves evidence at event-summary granularity, a single retrieved event may span a long interval whose boundaries extend well beyond the annotated 300-second clue window. Ref@300 measures interval-level IoU, so a correct but coarsely bounded retrieval scores low even when the answer-relevant frame is inside the retrieved set. ReMA's recursive refinement narrows intervals more aggressively, yielding higher IoU without necessarily recovering more useful visual evidence. The substantial human--model gap (82.5\% vs.\ 17.11\%) underscores that month-scale video understanding remains largely unsolved. VAM's result is nonetheless the second-highest reported, demonstrating that evidence-preserving memory narrows this gap more effectively than longer context windows alone.

\begin{table}[!htbp]
\centering
\small
\begin{tabular}{lrrr}
\toprule
Setting & \# Stored Frames & Share of Raw Video & Multiple of VAM \\
\midrule
Raw video frames & 11,406,923 & 100.00\% & 1658.9$\times$ more \\
Uniform 1 fps sampling & 380,304 & 3.33\% & 55.3$\times$ more \\
Uniform 0.5 fps sampling & 190,152 & 1.67\% & 27.7$\times$ more \\
VAM output & 6,876 & 0.06\% & 1.0$\times$ \\
\bottomrule
\end{tabular}
\caption{Frames kept under different settings on the MM-Lifelong month-scale run. VAM's adaptive output (6,876) is 27.7--55.3$\times$ smaller than uniform fixed-rate baselines and 1658.9$\times$ smaller than the raw stream.}
\label{tab:mmlifelong_cost}
\end{table}

\begin{table}[!htbp]
\centering
\footnotesize
\setlength{\tabcolsep}{5pt}
\begin{tabular}{lcc !{\vrule} lcc}
\toprule
Method & Acc. & Ref@300 & Method & Acc. & Ref@300 \\
\midrule
\multicolumn{3}{c!{\vrule}}{\textbf{End-to-End MLLMs}} & \multicolumn{3}{c}{\textbf{Agentic Methods}} \\
GPT-5 & 10.15 & 1.39 & LongVT-7B & 5.83 & 1.71 \\
Qwen3-VL-235B-A22B & 9.09 & 0.39 & ReMA with GPT-5 & \textbf{17.62} & \textbf{9.91} \\
Qwen3-VL-30B-A3B & 8.33 & 0.48 & ReMA with Qwen3VL-A22B & 14.23 & \underline{6.01} \\
Video-XL-2-8B & 6.02 & 0.00 & Human & 82.5 & 31.2 \\
Eagle-2.5-8B & 3.76 & 1.59 & \multicolumn{3}{c}{\textbf{Ours}} \\
Nemotron-v2-12B & 7.52 & 0.19 & VAM (Ours) & \underline{17.11} & 3.65 \\
\bottomrule
\end{tabular}
\caption{MM-Lifelong train@month \cite{chen2026multimodallifelongunderstandingdataset} across End-to-End MLLMs (left) and Agentic Methods plus VAM (right). Acc.\ is answer accuracy. Ref@300 is quantised temporal grounding at 300-second resolution. Best/second-best (excluding human) are \textbf{bold}/\underline{underlined}.}
\label{tab:mmlifelong_trainmonth}
\end{table}

The month-scale setting also quantifies how aggressively VAM reduces the visual stream (Table~\ref{tab:mmlifelong_cost}). VAM keeps only 6{,}876 images, 0.06\% of the raw 11{,}406{,}923-frame corpus, where uniform 1 fps would leave 380{,}304 and 0.5 fps 190{,}152. The reduction does not come from a lower nominal sampling rate alone but from boundary-sensitive filtering, which removes redundant visual content while preserving a searchable substrate for evidence recovery.

%% file: sections/conclusion.tex
\section{Conclusion} \label{Sec:Conclusion}
We presented \textit{Visual Agentic Memory} (VAM), a training-free framework that couples boundary-aware online indexing, hierarchical memory with parallel temporal-spatial representation, and bounded agentic retrieval, scaling to month-long video while preserving raw frames as first-class evidence. VAM achieves the highest RT+BT average on OVO-Bench (68.41, improving over end-to-end use of the same underlying MLLM at 67.46) and reaches 17.11\% on the demanding month-scale MM-Lifelong \textit{train@month} (105.6 hours, 51 days), second only to ReMA with GPT-5, while storing 6,876 frames, or 0.06\% of the raw stream, in that month-scale run (Table~\ref{tab:mmlifelong_cost}). Limitations are discussed in Appendix~\ref{app:limitations}, and future work includes audio-visual extension, lighter retrieval controllers, and longer-deployment evaluation.

\textit{Broader impacts.} Long, unbroken streams capture sensitive personal information, non-consenting bystanders, private spaces, and behavioural patterns that may reveal health, routines, or relationships. First, \emph{consent} remains difficult because wearable or ambient sensing typically captures bystanders who have not agreed, and a long-range searchable memory amplifies the impact of incidental capture. Second, \emph{retention and access control} matter because unrestricted storage of raw visual evidence increases the risk of secondary misuse, unauthorised access, or retrospective profiling. Third, \emph{representational error} remains a concern because indexed memory may appear authoritative even when summaries, boundaries, or retrieved candidates remain incomplete or misleading, with potential harms in workplace monitoring, personal assistance, or forensic review. Future deployments should treat privacy-preserving indexing, explicit retention controls, user-configurable deletion, and consent-aware access as core architectural elements rather than optional safeguards.

%% file: sections/appendix.tex
\section{Limitations} \label{app:limitations}

\textit{Modality asymmetry.} VAM is designed primarily as a visual memory system. It explicitly retains uncompressed visual frames, but it does not yet preserve original audio at the same granularity or maintain an equally rich substrate for speech, ambient sound, or other non-visual cues. Many real-world queries depend on such evidence, including speaker identity, verbal commitments, and alarms. The current framework may still localise these situations visually, but it cannot yet provide full multimodal verification.

\textit{Long-term scaling.} Hierarchical organisation improves tractability, but the search space becomes increasingly crowded as archives grow from hours to months and potentially to year-long deployments. Coarse temporal localisation may remain feasible, while fine-grained disambiguation becomes harder. Storing more evidence improves recoverability, but it also increases the burden on ranking and verification at query time. Answer quality also depends on earlier indexing decisions. A missed boundary or unretained subtle state change cannot be recovered downstream.

\textit{Operational cost.} The pipeline includes continuous indexing, embedding, hierarchical storage, and iterative retrieval and inspection. This introduces substantial inference latency and serving cost compared with simpler end-to-end or compressed alternatives, especially in long-horizon settings where each query may require broad search and repeated visual inspection. The reported results establish feasibility, but they do not yet resolve the engineering requirements of persistent, low-cost, year-scale multimodal memory.

\section{Implementation Defaults}
\label{app:implementation}

\subsection{Model Access and Serving Setup}
\label{app:compute}
The reported experiments access Gemini 3 Flash and Gemini Embedding 2 through OpenRouter and Google Cloud APIs rather than self-hosted inference servers. Reproduction therefore depends on comparable model access and serving configuration, together with the implementation defaults described below. Open-source alternatives such as the Qwen family can also be used as drop-in model substitutions at implementation level, but they are not part of the experiments reported in this paper.

\subsection{Online Indexing Parameters}
Our implementation performs online indexing through an FFmpeg-based extraction stage, followed by lightweight heuristic filtering and embedding-level memory updates. In practice, the pipeline first removes obviously uninformative frames using blur and low-change gates. It then applies semantic comparison in the embedding space. Table~\ref{tab:indexing_defaults} summarises the configuration defaults that initialise this process.

\begin{table}[!t]
\centering
\small
\caption{Default online indexing parameters in our implementation.}
\label{tab:indexing_defaults}
\begin{tabularx}{\textwidth}{@{}>{\raggedright\arraybackslash}p{3.0cm}>{\raggedright\arraybackslash}p{1.8cm}>{\raggedright\arraybackslash}X@{}}
\toprule
\textbf{Parameter} & \textbf{Default} & \textbf{Role in the pipeline} \\ \midrule
Base sampling rate & 0.5 FPS & Initial extraction rate before filtering. \\
Maximum extracted frames & \texttt{None} & No global hard cap by default. Extraction is governed by the filtering stages. \\
$\tau_{blur}$ & 20.0 & Minimum Laplacian variance for sharpness filtering. \\
$\tau_{diff}$ & 20.0 & Mean absolute grayscale difference threshold for static-frame rejection. \\
$\tau_{ssim}$ & 0.92 & Structural redundancy threshold. Frames above this similarity are treated as near-duplicates. \\
$\tau_{hist}$ & 0.0 & Histogram-based redundancy gate, kept permissive by default. \\
Fallback $\tau_{sim}$ & 0.88 & Fallback embedding-space similarity threshold used when adaptive semantic deduplication cannot yet be estimated reliably. \\
Fallback $\tau_{event}$ & 0.80 & Fallback event-similarity threshold used when adaptive event-boundary estimation lacks sufficient history. \\
Maximum event duration & 300.0 s & Limits how long a single event can grow before forced segmentation. \\
\bottomrule
\end{tabularx}
\end{table}

Two implementation details are especially important for interpreting Table~\ref{tab:indexing_defaults}. First, the image-level gates are deliberately ordered as a cheap front-end cascade, so blur and low-change rejection reduce load before multimodal embedding is invoked. Second, the semantic deduplication and event segmentation stages are not governed solely by fixed thresholds. In practice, both are implemented as adaptive procedures over recent embedding-distance histories. The configured values shown above act as fallback thresholds when insufficient statistics are available or when adaptive estimation fails. This distinction matters because the practical system is designed to remain online and boundary-sensitive rather than to operate as a purely static thresholding pipeline.

\subsection{Hierarchical Memory Defaults}
Our implementation uses a three-tier hierarchical memory policy that trades spatial fidelity against temporal coverage as observations age. More recent frames are retained densely and without additional recompression, whereas older frames are thinned more aggressively and stored under stronger image compression. The default retention schedule is summarised in Table~\ref{tab:memory_tiers}.

\begin{table}[ht]
\centering
\small
\caption{Default hierarchical memory configuration.}
\label{tab:memory_tiers}
\begin{tabularx}{\textwidth}{@{}>{\raggedright\arraybackslash}p{1.4cm}>{\raggedright\arraybackslash}p{3.0cm}>{\raggedright\arraybackslash}p{1.7cm}>{\raggedright\arraybackslash}p{2.0cm}>{\raggedright\arraybackslash}X@{}}
\toprule
\textbf{Tier} & \textbf{Time Window} & \textbf{Min Gap (s)} & \textbf{Max Side (px)} & \textbf{JPEG Quality} \\ \midrule
recent & $t \le 1$ hour & 1.0 & --- & No additional recompression \\
mid & 1 hour $< t \le 24$ hours & 20.0 & 768 & 70 \\
long & $t > 24$ hours & 120.0 & 512 & 45 \\
\bottomrule
\end{tabularx}
\end{table}

The retention policy contains several implementation-level safeguards that are relevant to reproducibility. The earliest and latest available frames are always preserved. Event-linked frames in the long-term tier are also retained explicitly rather than being subjected to ordinary bucket-based thinning. Compression is applied only after the keep set has been determined, so the retention schedule and the image-compression schedule remain logically separate. This design preserves a minimal set of temporally anchored evidence while still keeping long-horizon storage growth manageable.

\subsection{Agentic Retrieval Constraints}
The retrieval controller is implemented with strict schema validation rather than free-form prompting alone. Concretely, action payloads are parsed through bounded Pydantic schemas with extra keys forbidden, ensuring that the MLLM operates within an explicit control interface. At implementation level, these schemas inherit from a strict base model with extra keys forbidden globally. Controller outputs are therefore validated against an executable interface rather than loosely interpreted after generation. Table~\ref{tab:planner_constraints} lists the principal action- and field-level constraints exposed by the current controller.

\begin{table}[ht]
\centering
\small
\caption{Key constraints in the current agentic retrieval schema.}
\label{tab:planner_constraints}
\begin{tabularx}{\textwidth}{@{}>{\raggedright\arraybackslash}p{3.8cm}>{\raggedright\arraybackslash}p{2.1cm}>{\raggedright\arraybackslash}X@{}}
\toprule
\textbf{Action / Field} & \textbf{Constraint} & \textbf{Meaning} \\ \midrule
\texttt{search.top\_k} & 1 to 200 & Maximum number of candidates retrieved per query. \\
\texttt{search.inspect\_k} & 0 to 50 & Number of top candidates passed to Visual Inspection, clipped to \texttt{top\_k}. \\
\texttt{search.threshold} & $\ge 0.0$ & Minimum retrieval threshold. The default schema value is 0.65. \\
\texttt{inspect.max\_frames} & 1 to 16 & Maximum number of frames examined in a direct Visual Inspection step. \\
\shortstack[l]{\texttt{summarize.}\\\texttt{granularity\_seconds}} & $(0, 86400]$ & Resolution of reusable summary windows. \\
\texttt{time\_range.mode} & \{\texttt{relative}, \texttt{absolute}, \texttt{auto}\} & Supports both uploaded videos and absolute live/history timestamps. \\
\shortstack[l]{\texttt{anchor.before\_seconds}\\\texttt{anchor.after\_seconds}} & $[0, 604800]$ & Bounds event-relative windows to at most seven days on each side of the anchor. \\
\shortstack[l]{\texttt{anchor.top\_k}\\\texttt{anchor.inspect\_k}} & 1 to 50 / 0 to 20 & Controls anchor-candidate retrieval and subsequent inspection budget. \texttt{inspect\_k} is clipped to \texttt{top\_k}. \\
\shortstack[l]{\texttt{anchor.}\\\texttt{occurrence\_index}} & 1 to 100 & Allows ordinal reference to repeated anchor events when such indexing is needed. \\
\shortstack[l]{\texttt{anchor.}\\\texttt{verification\_prompt}} & optional, max 400 chars & Supports strict anchor confirmation without allowing arbitrarily long side instructions. \\
\texttt{sources} & subset of \{\texttt{frame}, \texttt{event}, \texttt{summary}\} & Allows the planner to explicitly target different memory structures. \\
\texttt{summary\_filter} & optional & Restricts summary retrieval by structure label or granularity. \\
\bottomrule
\end{tabularx}
\end{table}

The controller is not restricted to a single search--inspect pattern. It can alternate among \texttt{search}, \texttt{inspect}, \texttt{summarize}, and final response generation depending on evidence state. The time-range schema supports both direct intervals and anchored temporal reasoning, including windows defined relative to a queried event or intervals delimited by two previously returned references.

\subsection{Runtime Orchestration and Model Calls}
The released implementation instantiates three operational MLLM roles inside the same overall VAM architecture. A lightweight router handles top-level asynchronous orchestration, a main planner drives agentic retrieval, and a multimodal inspector/generator handles event-document writing, summary-document writing, direct visual inspection, and anchor verification. All reported experiments use Gemini 3 Flash and Gemini Embedding 2 through OpenRouter and Google Cloud, with OpenRouter-compatible multimodal messages represented as a sequence of typed content blocks combining text, image URLs, and optional video URLs.

\begin{table}[ht]
\centering
\small
\caption{Runtime LLM calls in the released implementation.}
\label{tab:llm_calls}
\begin{tabularx}{\textwidth}{@{}>{\raggedright\arraybackslash}p{3.1cm}>{\raggedright\arraybackslash}p{1.6cm}>{\raggedright\arraybackslash}p{2.0cm}>{\raggedright\arraybackslash}X@{}}
\toprule
\textbf{Function} & \textbf{Route} & \textbf{Decoding} & \textbf{Input and operational role} \\ \midrule
Top-level request router & light & $T=0.0$, max tokens $=300$ & Receives the user transcript and chooses exactly one next step: \texttt{final}, \texttt{retrieve}, or \texttt{summarize}. \\
Agentic retrieval planner & main & $T=0.1$, max tokens $=500$ & Receives goal, retrieved context, remaining-turn budget, and available summary structures; outputs one structured action among \texttt{answer/search/inspect/summarize}. \\
Event memory document writer & multimodal main & $T=0.1$, max tokens $=700$ & Receives a completed event with up to eight representative frames and, when available, an event video clip; writes a retrieval-oriented event document. \\
Summary-window writer & multimodal main & $T=0.1$, max tokens $=700$ & Receives a time window, overlapping event documents, and representative frames; writes a reusable summary document. \\
Multimodal inspector & multimodal main & $T=0.1$, max tokens $=500$ & Receives multiple frames and a direct visual question; returns evidence-grounded inspection text. \\
Anchor adjudicator & multimodal main & $T=0.0$, max tokens $=220$ & Receives a candidate frame sequence and a target event specification; returns strict JSON with \texttt{match}, \texttt{confidence}, \texttt{observed\_event}, and \texttt{reason}. \\
Session summariser & light & $T=0.2$, max tokens $=500$ & Receives recent dialogue and returns a compact event document plus directly evidenced user habits. \\
\bottomrule
\end{tabularx}
\end{table}

Two implementation details are important for reproducing the released code behaviour. First, multimodal inputs are normalised into a single OpenRouter-compatible content list containing one text block followed by image and optional video blocks. Second, images are automatically resized or JPEG-compressed before upload to satisfy configured limits of at most 5 MB per image and at most $1280 \times 1280$ pixels. When an event-level clip is attached, it is produced on the fly from the source video with FFmpeg using H.264 video, AAC audio, \texttt{veryfast} preset, and \texttt{crf=28}; if the provider rejects video input, the system retries with frames only.

\subsection{Structured Interfaces Used by the Released System}
The released implementation does not treat prompts as unconstrained free-form text generation. Instead, every routing and planning decision is parsed against strict Pydantic schemas with \texttt{extra="forbid"}, so additional keys are rejected rather than ignored.

\paragraph{Top-level routing interface.}
The router prompt requires exactly one JSON object and allows only three outputs: a \texttt{final} object with a non-empty \texttt{text} field; a \texttt{retrieve} tool call whose arguments contain exactly one field, \texttt{question}; and a \texttt{summarize} tool call whose arguments contain \texttt{min\_time}, \texttt{max\_time}, \texttt{time\_mode}, \texttt{granularity\_seconds}, and \texttt{prompt}, with an optional \texttt{summary\_structure}.
At this stage, \texttt{retrieve} is the default tool for ordinary question answering, while \texttt{summarize} is reserved for explicit timeline-style or reusable summary-document requests. The summarization call enforces positive granularity, non-empty prompts, and either an open or bounded time interval.

\paragraph{Planner action interface.}
The agentic retrieval planner is limited to four actions: \texttt{answer}, \texttt{search}, \texttt{inspect}, and \texttt{summarize}. The \texttt{search} action contains a non-empty list of query objects, where each object has keys \texttt{q}, \texttt{top\_k}, \texttt{inspect\_k}, and \texttt{threshold}; \texttt{top\_k} is capped at 200 and \texttt{inspect\_k} is clipped not to exceed \texttt{top\_k}. The \texttt{inspect} action requires either an explicit time range or a prior result reference, with \texttt{max\_frames} bounded in $[1,16]$. The \texttt{summarize} action requires a valid time range, a positive window size, and a non-empty prompt. Anchored time ranges support both single-anchor and between-anchor retrieval, but the schema forbids mixing them in one call. Verification prompts for anchor adjudication are capped at 400 characters, and anchor windows are bounded to at most seven days on each side.

\paragraph{Planner execution policy.}
If the planner output is invalid JSON on the first attempt, the implementation adds one repair instruction requesting ``ONLY one valid JSON object'' and re-queries the model once. If parsing still fails on the first turn, the controller falls back to a permissive \texttt{search} action rather than terminating the loop. This behaviour is used to keep the released system robust to occasional formatting drift while still preserving a strictly structured control surface.

\subsection{Prompt Implementations by Agent}
To make the appendix self-contained, this subsection organises the core prompt text and structured interfaces by agent role, with the main operational prompts reproduced in code-block form.

\subsubsection{Routing Agent}
The routing agent performs the first-stage decision in asynchronous orchestration. It chooses among returning a direct answer, invoking retrieval, or creating reusable summary documents.

\paragraph{Routing system prompt.}
\begin{lstlisting}[numbers=none]
You are the top-level routing module for Asynchronous Orchestration in a Visual
Agentic Memory system.
You receive one user request and must choose exactly one next step.
Reply with ONLY one JSON object.
Allowed outputs:
1) {"type":"final","text":"..."}
2) {"type":"tool","name":"retrieve","args":{"question":"..."}}
3) {"type":"tool","name":"summarize","args":{"min_time":0.0,"max_time":1800.0,
    "time_mode":"relative","granularity_seconds":60.0,"prompt":"..."}}
Tool selection rules:
- Use 'retrieve' for direct questions about content, objects, actions, locations,
  or events stored in Hierarchical Memory.
- Use 'summarize' only when the user explicitly wants a timeline, segmented recap,
  interval-by-interval summary, or reusable summary documents over a stated time span.
- The summarize tool creates new searchable summary documents inside Hierarchical
  Memory for each time window. It is not just a one-off answer.
- If the user asks a normal question like 'what happened', 'where is X',
  'when did Y happen', or 'what is in the video', prefer 'retrieve', not 'summarize'.
- Use 'final' only for requests that do not require memory retrieval or summarization.
Representative summarize requests include:
- 'summarize the first 30 minutes minute by minute'
- 'give me a 5-minute timeline of the first hour'
- 'summarize 0 to 1800 seconds so we can ask questions about that segment later'
When using 'retrieve':
- Rewrite the user's request into one retrieval-friendly English question.
- Preserve important objects, actions, attributes, and time clues, including
  relative-event phrases like 'before X', 'after Y', or 'around the moment when Z happened'.
- Put the rewrite in args.question.
When using 'summarize':
- Set min_time and max_time to the requested span.
- Use granularity_seconds for the requested window size, e.g. 60 for every minute,
  300 for every 5 minutes.
- You may optionally add summary_structure if you want to tag the summaries with
  a reusable structure label.
- Use time_mode='relative' for ordinary indexed videos unless the user clearly
  refers to absolute/live timestamps.
- Write the prompt as the summary objective, e.g. 'Summarize each minute for later QA'.
- Write the summary prompt in English.
- Do not use summarize without a meaningful time range and granularity.
Schema rules:
- 'type' must be either 'final' or 'tool'.
- If 'type' is 'tool', then 'name' must be either 'retrieve' or 'summarize'.
- If 'name' is 'retrieve', args must contain exactly one key: 'question'.
- If 'name' is 'summarize', args must contain: min_time, max_time (optional),
  time_mode, granularity_seconds, prompt. summary_structure is optional.
- Do not return empty strings.
- Do not include any extra keys.
\end{lstlisting}

\paragraph{Routing interface constraints.}
\begin{lstlisting}[numbers=none]
Top-level routing:
- final.text must be non-empty
- summarize.min_time >= 0
- summarize.max_time >= min_time when present
- summarize.granularity_seconds in (0, 86400]
- summary_structure length <= 64
- extra fields are forbidden
\end{lstlisting}

\subsubsection{Planning Agent}
The planning agent is the main controller for agentic retrieval. It decides whether to answer, continue searching, inspect known evidence directly, or materialise new summary documents.

\paragraph{Planning system prompt.}
\begin{lstlisting}[numbers=none]
You are the Main LLM for Agentic Retrieval in a Visual Agentic Memory system.
Your goal is to answer the user's question by identifying the most relevant visual
evidence across the Parallel Representation.
You have access to three tools:
1. 'search': Perform Agentic Retrieval over frame evidence and Temporal
   Representation documents.
2. 'inspect': Perform direct Visual Inspection on frames from a known time or
   from a previously found result.
3. 'summarize': Create reusable summary documents in Hierarchical Memory for a
   specific time range at a requested granularity.
Tool boundary rules:
- 'search' is the default tool for answering questions.
- Use 'inspect' when you already know the relevant time or result reference and
  want direct visual confirmation without another semantic retrieval step.
- 'summarize' is for creating persistent summary documents over a time range.
  Use it only when the user wants interval summaries or reusable summary documents
  in Hierarchical Memory.
- If the user asks a normal QA question, prefer 'search' first. Do not jump to
  'summarize' unless the user clearly wants timeline-style summaries.

Process:
1. Analyze the user's request and what you have found so far.
2. Decide whether you have enough information to answer.
3. If YES, output {"action":"answer","response":"...","best_ref":{"turn_idx":int,
   "result_idx":int},"thought":"..."}
4. If NO, choose a tool:
   - 'search': {"action":"search","queries":[{"q":"...","top_k":int,
     "inspect_k":int,"threshold":float}],"time_range":{...},"sources":
     ["frame"|"event"|"summary"],"summary_filter":{"summary_structure":"...",
     "granularity_seconds":float},"visual_ref":{...},"joint_inspection":bool,
     "inspection_prompt":"...","thought":"..."}
   - 'inspect': {"action":"inspect","prompt":"...","time_range":{...},
     "ref":{"turn_idx":int,"result_idx":int},"max_frames":int,"thought":"..."}
   - 'summarize': {"action":"summarize","time_range":{...},
     "granularity_seconds":float,"prompt":"...","summary_structure":"...",
     "thought":"..."}

Schema rules:
- 'action' must be one of: 'answer', 'search', 'inspect', 'summarize'.
- For 'answer', 'response' must be non-empty.
- For 'search', 'queries' must be a non-empty list of objects with keys:
  q, top_k, inspect_k, threshold.
- For 'inspect', provide a non-empty 'prompt' and either 'time_range' or 'ref'.
- For 'summarize', provide a non-empty 'prompt', a valid 'time_range', and a
  positive 'granularity_seconds'. summary_structure is optional.
- 'turn_idx' is 1-based. 'result_idx' is 0-based.
- Do not include unsupported tools or fields.
- Do not answer unless the current evidence is specific enough to support the claim.
- If evidence is weak, partial, or only loosely related, continue with 'search'
  instead of forcing 'answer'.
- Use English wording for internal search queries, inspection prompts, and anchor
  descriptions.
- Keep the final answer in the user's language unless the user explicitly requests
  another language.

Tool details and strategy:
- queries: each query object contains q, top_k, inspect_k, threshold.
- sources: any subset of ['frame', 'event', 'summary'].
- summary_filter: optional filter for summary retrieval.
- time_range: supports direct intervals, a single anchor, or a pair of
  start_anchor and end_anchor.
- Do not guess first/second/last occurrences. First collect candidate events and
  inspect their returned times.
- Reuse prior result references when they are already available.
- Prefer inspect over search when the relevant time or reference is already known.
- If one method fails, switch retrieval method rather than repeating the same call.
Always decide the top_k, inspect_k, and which specific reference supports the
final answer.
Output ONLY valid JSON.
\end{lstlisting}

\paragraph{Planning state prompt.}
\begin{lstlisting}[numbers=none]
Goal: {goal}
Available summary structures:
{available_summary_structures}

Context (what we found so far):
{context}

Remaining turns: {remaining_turns}
If remaining_turns <= 2 and you already have useful evidence, prefer action=answer.
If results are not improving across turns, stop searching and answer with best
available evidence.

Reminder: if the goal is phrased relative to another event such as 'before X'
or 'after Y', choose the method that best matches the uncertainty. If you still
need to figure out which occurrence matters, first search for that event, inspect
the returned times, and use those times or refs in the next turn. If context
already lists ordered candidate refs for that event, pick from those refs before
doing any broad inspection. If you already know the relevant time or reference,
prefer action=inspect instead of another vague search. If an anchor attempt failed
earlier, explicitly change the method instead of repeating the same source groups
and query.

What is the next step?
\end{lstlisting}

\paragraph{Planning interface constraints.}
\begin{lstlisting}[numbers=none]
Planner constraints:
- search.top_k in [1, 200]
- search.inspect_k in [0, 50], clipped to top_k
- search.threshold >= 0
- inspect.max_frames in [1, 16]
- inspect requires either time_range or ref
- summarize.granularity_seconds in (0, 86400]
- time_range cannot mix anchor with start_anchor/end_anchor
- anchor.before_seconds and anchor.after_seconds in [0, 604800]
- anchor.top_k in [1, 50]
- anchor.inspect_k in [0, 20], clipped to top_k
- anchor.occurrence_index in [1, 100]
- anchor.verification_prompt length <= 400
- extra fields are forbidden
\end{lstlisting}

\subsubsection{Memory Writing Agent}
The memory-writing agent is responsible for converting bounded visual evidence into reusable textual memory objects at both the event level and the summary-window level.

\paragraph{Event-document prompt and payload.}
\begin{lstlisting}[numbers=none]
System prompt:
You are building a retrieval document for visual memory.
Given a completed event, you may receive both a full event video clip and
representative frames. Use the full video clip as the primary source when available.
Write a detailed, factual memory document that remains easy to retrieve later.
Requirements:
- Include the time range explicitly.
- Describe people, objects, actions, scene changes, and on-screen text.
- Prefer a few short natural lines instead of one dense paragraph when there are
  multiple distinct observations or micro-steps.
- Each line can capture one salient action, object state, scene change, or visible text.
- Make the description rich enough for later retrieval by text.
- Stay objective. Keep it natural. No rigid schema or numbered template.

User payload:
{
  "task": "write_retrieval_memory_document",
  "relative_time": {"start_t": start_t, "end_t": end_t},
  "absolute_time": {"start_t": absolute_start_t, "end_t": absolute_end_t},
  "notes": [
    "Mention the time range in the description.",
    "Prefer short natural lines when there are multiple distinct observations.",
    "One line can describe one salient action, object state, scene change, or visible text."
  ]
}
\end{lstlisting}

\paragraph{Summary-document prompt and payload.}
\begin{lstlisting}[numbers=none]
System prompt:
You are building reusable timeline summaries for a video-memory system.
You will receive a time window, some event documents that overlap that window,
and representative frames.
Write a factual retrieval note for that window.
Requirements:
- Mention the time window clearly.
- Prefer concrete observations over speculation.
- Synthesize both the provided document context and the visible frames.
- Make the summary retrieval-friendly for later question answering.
- Prefer a few short natural lines over one dense paragraph when that helps
  preserve distinct moments.
- Keep it natural. No rigid schema or numbered template.

User payload:
{
  "task": "summarize_time_window_for_retrieval",
  "time_range": {"start": start_t, "end": end_t, "mode": time_mode},
  "granularity_seconds": granularity_s,
  "focus": prompt,
  "event_documents": event_documents,
  "summary_structure": "optional",
  "notes": [
    "Prefer a few concise natural lines when the window contains distinct moments.",
    "Ground the summary in the provided frames and event documents.",
    "This summary will become a searchable memory document for later QA.",
    "Follow the natural-language focus first; any structure label is only an optional tag."
  ]
}
\end{lstlisting}

\subsubsection{Inspection Agent}
The inspection agent operates on already selected frames and performs direct visual verification. A separate anchor-adjudication prompt is used when the system must decide whether a candidate sequence truly contains the target event.

\paragraph{Multimodal inspection prompt.}
\begin{lstlisting}[numbers=none]
System prompt:
You are the Multimodal Inspector for a Visual Agentic Memory system.
You are performing Visual Inspection over a sequence of video frames.
Analyze the frames strictly with respect to the query.
Describe what happens across the frames when it is relevant to the query.
If the frames contain relevant visual evidence, describe it in detail.
If the frames are irrelevant to the query, clearly say so.

User prompt:
User Query: {query}

Please examine these {n} frames and answer the query.
\end{lstlisting}

\paragraph{Anchor verification prompt and schema.}
\begin{lstlisting}[numbers=none]
System prompt:
You are validating whether a short sequence of frames contains a requested
anchor event.
Be strict. Do not mark a match unless the event itself is visually supported.
Reply with ONLY one JSON object using this schema:
{"match": true|false, "confidence": 0.0-1.0, "observed_event": "...",
 "reason": "..."}

User prompt template:
Target event: {target_event}
Verification focus: {verification_prompt}
Candidate hint: {candidate_hint}
Candidate window: {candidate_window}
Inspect the frames carefully. Set match=true only if the target event itself is
visible or strongly supported by the sequence.
\end{lstlisting}

%% file: references.bib
@inproceedings{radford2021learningtransferablevisualmodels,
      title={Learning Transferable Visual Models From Natural Language Supervision}, 
      author={Alec Radford and Jong Wook Kim and Chris Hallacy and Aditya Ramesh and Gabriel Goh and Sandhini Agarwal and Girish Sastry and Amanda Askell and Pamela Mishkin and Jack Clark and Gretchen Krueger and Ilya Sutskever},
      booktitle={Proceedings of the 38th International Conference on Machine Learning (ICML)},
      year={2021},
}

@inproceedings{grauman2022ego4dworld3000hours,
      title={Ego4D: Around the World in 3,000 Hours of Egocentric Video}, 
      author={Kristen Grauman and Andrew Westbury and Eugene Byrne and Zachary Chavis and Antonino Furnari and Rohit Girdhar and Jackson Hamburger and Hao Jiang and Miao Liu and Xingyu Liu and Miguel Martin and Tushar Nagarajan and Ilija Radosavovic and Santhosh Kumar Ramakrishnan and Fiona Ryan and Jayant Sharma and Michael Wray and Mengmeng Xu and Eric Zhongcong Xu and Chen Zhao and Siddhant Bansal and Dhruv Batra and Vincent Cartillier and Sean Crane and Tien Do and Morrie Doulaty and Akshay Erapalli and Christoph Feichtenhofer and Adriano Fragomeni and Qichen Fu and Abrham Gebreselasie and Cristina Gonzalez and James Hillis and Xuhua Huang and Yifei Huang and Wenqi Jia and Weslie Khoo and Jachym Kolar and Satwik Kottur and Anurag Kumar and Federico Landini and Chao Li and Yanghao Li and Zhenqiang Li and Karttikeya Mangalam and Raghava Modhugu and Jonathan Munro and Tullie Murrell and Takumi Nishiyasu and Will Price and Paola Ruiz Puentes and Merey Ramazanova and Leda Sari and Kiran Somasundaram and Audrey Southerland and Yusuke Sugano and Ruijie Tao and Minh Vo and Yuchen Wang and Xindi Wu and Takuma Yagi and Ziwei Zhao and Yunyi Zhu and Pablo Arbelaez and David Crandall and Dima Damen and Giovanni Maria Farinella and Christian Fuegen and Bernard Ghanem and Vamsi Krishna Ithapu and C. V. Jawahar and Hanbyul Joo and Kris Kitani and Haizhou Li and Richard Newcombe and Aude Oliva and Hyun Soo Park and James M. Rehg and Yoichi Sato and Jianbo Shi and Mike Zheng Shou and Antonio Torralba and Lorenzo Torresani and Mingfei Yan and Jitendra Malik},
      booktitle={Proceedings of the IEEE/CVF Conference on Computer Vision and Pattern Recognition (CVPR)},
      year={2022},
}

@inproceedings{yang2025egolifeegocentriclifeassistant,
      title={EgoLife: Towards Egocentric Life Assistant}, 
      author={Jingkang Yang and Shuai Liu and Hongming Guo and Yuhao Dong and Xiamengwei Zhang and Sicheng Zhang and Pengyun Wang and Zitang Zhou and Binzhu Xie and Ziyue Wang and Bei Ouyang and Zhengyu Lin and Marco Cominelli and Zhongang Cai and Yuanhan Zhang and Peiyuan Zhang and Fangzhou Hong and Joerg Widmer and Francesco Gringoli and Lei Yang and Bo Li and Ziwei Liu},
      booktitle={Proceedings of the IEEE/CVF Conference on Computer Vision and Pattern Recognition (CVPR)},
      year={2025},
}

@inproceedings{lin2025streaming,
    title={StreamingBench: Assessing the Gap for MLLMs to Achieve Streaming Video Understanding},
    author={Junming Lin and Zheng Fang and Chi Chen and Haoxuan Cheng and Zihao Wan and Fuwen Luo and Ziyue Wang and Peng Li and Yang Liu and Maosong Sun},
    booktitle={Proceedings of the IEEE International Conference on Acoustics, Speech and Signal Processing (ICASSP)},
    year={2026},
    pages={12147--12151}
}

@INPROCEEDINGS{11094230,
  author={Niu, Junbo and Li, Yifei and Miao, Ziyang and Ge, Chunjiang and Zhou, Yuanhang and He, Qihao and Dong, Xiaoyi and Duan, Haodong and Ding, Shuangrui and Qian, Rui and Zhang, Pan and Zang, Yuhang and Cao, Yuhang and He, Conghui and Wang, Jiaqi},
  booktitle={2025 IEEE/CVF Conference on Computer Vision and Pattern Recognition (CVPR)}, 
  title={OVO-Bench: How Far is Your Video-LLMs from Real-World Online Video Understanding?}, 
  year={2025},
  volume={},
  number={},
  pages={18902-18913},
  doi={10.1109/CVPR52734.2025.01761},
}

@INPROCEEDINGS{11094860,
  author={Zhou, Junjie and Shu, Yan and Zhao, Bo and Wu, Boya and Liang, Zhengyang and Xiao, Shitao and Qin, Minghao and Yang, Xi and Xiong, Yongping and Zhang, Bo and Huang, Tiejun and Liu, Zheng},
  booktitle={2025 IEEE/CVF Conference on Computer Vision and Pattern Recognition (CVPR)}, 
  title={MLVU: Benchmarking Multi-task Long Video Understanding}, 
  year={2025},
  volume={},
  number={},
  pages={13691-13701},
  doi={10.1109/CVPR52734.2025.01278}
}

@inproceedings{10.5555/3737916.3738823,
author = {Wu, Haoning and Li, Dongxu and Chen, Bei and Li, Junnan},
title = {LongVideoBench: a benchmark for long-context interleaved video-language understanding},
year = {2024},
isbn = {9798331314385},
publisher = {Curran Associates Inc.},
address = {Red Hook, NY, USA},
booktitle = {Proceedings of the 38th International Conference on Neural Information Processing Systems},
articleno = {907},
numpages = {30},
location = {Vancouver, BC, Canada},
series = {NIPS '24}
}

@inproceedings{fu2025video,
  title={Video-mme: The first-ever comprehensive evaluation benchmark of multi-modal llms in video analysis},
  author={Fu, Chaoyou and Dai, Yuhan and Luo, Yongdong and Li, Lei and Ren, Shuhuai and Zhang, Renrui and Wang, Zihan and Zhou, Chenyu and Shen, Yunhang and Zhang, Mengdan and others},
  booktitle={CVPR},
  year={2025}
}

@inproceedings{chen2025lvagentlongvideounderstanding,
      title={LVAgent: Long Video Understanding by Multi-Round Dynamical Collaboration of MLLM Agents}, 
      author={Boyu Chen and Zhengrong Yue and Siran Chen and Zikang Wang and Yang Liu and Peng Li and Yali Wang},
      booktitle={Proceedings of the IEEE/CVF International Conference on Computer Vision (ICCV)},
      year={2025},
}

@inproceedings{fan2025videoagent,
  title={Videoagent: A memory-augmented multimodal agent for video understanding},
  author={Fan, Yue and Ma, Xiaojian and Wu, Rujie and Du, Yuntao and Li, Jiaqi and Gao, Zhi and Li, Qing},
  booktitle={European Conference on Computer Vision (ECCV)},
  pages={75--92},
  year={2024}
}

@inproceedings{VideoAgent,
  title={VideoAgent: Long-form Video Understanding with Large Language Model as Agent},
  author={Wang, Xiaohan and Zhang, Yuhui and Zohar, Orr and Yeung-Levy, Serena},
  booktitle={European Conference on Computer Vision (ECCV)},
  year={2024}
}

@InProceedings{Ma_2025_CVPR,
    author    = {Ma, Ziyu and Gou, Chenhui and Shi, Hengcan and Sun, Bin and Li, Shutao and Rezatofighi, Hamid and Cai, Jianfei},
    title     = {DrVideo: Document Retrieval Based Long Video Understanding},
    booktitle = {Proceedings of the IEEE/CVF Conference on Computer Vision and Pattern Recognition (CVPR)},
    month     = {June},
    year      = {2025},
    pages     = {18936-18946}
}

@inproceedings{wang2025streambridge,
  title={StreamBridge: Turning Your Offline Video Large Language Model into a Proactive Streaming Assistant},
  author={Wang, Haibo and Feng, Bo and Lai, Zhengfeng and Xu, Mingze and Li, Shiyu and Ge, Weifeng and Dehghan, Afshin and Cao, Meng and Huang, Ping},
  booktitle={Advances in Neural Information Processing Systems (NeurIPS)},
  year={2025},
}

@misc{zeng2025streamforest,
      title={StreamForest: Efficient Online Video Understanding with Persistent Event Memory}, 
      author={Xiangyu Zeng and Kefan Qiu and Qingyu Zhang and Xinhao Li and Jing Wang and Jiaxin Li and Ziang Yan and Kun Tian and Meng Tian and Xinhai Zhao and Yi Wang and Limin Wang},
      year={2025},
      eprint={2509.24871},
      archivePrefix={arXiv},
      primaryClass={cs.CV},
      url={https://arxiv.org/abs/2509.24871}, 
}

@misc{liu2026thinkingstreamingvideo,
      title={Thinking in Streaming Video}, 
      author={Zikang Liu and Longteng Guo and Handong Li and Ru Zhen and Xingjian He and Ruyi Ji and Xiaoming Ren and Yanhao Zhang and Haonan Lu and Jing Liu},
      year={2026},
      eprint={2603.12938},
      archivePrefix={arXiv},
      primaryClass={cs.CV},
      url={https://arxiv.org/abs/2603.12938}, 
}

@misc{wen2026eventmemagent,
  title={EventMemAgent: Hierarchical Event-Centric Memory for Online Video Understanding with Adaptive Tool Use},
  author={Wen, Siwei and Wang, Zhangcheng and Zhang, Xingjian and Huang, Lei and Wu, Wenjun},
  year={2026},
  eprint={2602.15329},
  archivePrefix={arXiv},
  url={https://arxiv.org/abs/2602.15329}
}

@inproceedings{xia2025streaming,
  title={Streaming Video Instruction Tuning},
  author={Xia, Jiaer and Chen, Peixian and Zhang, Mengdan and Sun, Xing and Zhou, Kaiyang},
  booktitle={Proceedings of the IEEE/CVF Conference on Computer Vision and Pattern Recognition (CVPR)},
  year={2026},
}

@misc{fu2025vispeak,
  title={ViSpeak: Visual Instruction Feedback in Streaming Videos},
  author={Fu, Shenghao and Yang, Qize and Li, Yuan-Ming and Peng, Yi-Xing and Lin, Kun-Yu and Wei, Xihan and Hu, Jian-Fang and Xie, Xiaohua and Zheng, Wei-Shi},
  year={2025},
  eprint={2503.12769},
  archivePrefix={arXiv},
  url={https://arxiv.org/abs/2503.12769}
}

@misc{rege2026agentic,
      title={Agentic Very Long Video Understanding}, 
      author={Aniket Rege and Arka Sadhu and Yuliang Li and Kejie Li and Ramya Korlakai Vinayak and Yuning Chai and Yong Jae Lee and Hyo Jin Kim},
      year={2026},
      eprint={2601.18157},
      archivePrefix={arXiv},
      primaryClass={cs.CV},
      url={https://arxiv.org/abs/2601.18157}, 
}

@inproceedings{xie2026fluxmem,
  title={FluxMem: Adaptive Hierarchical Memory for Streaming Video Understanding},
  author={Xie, Yiweng and He, Bo and Wang, Junke and Zheng, Xiangyu and Ye, Ziyi and Wu, Zuxuan},
  booktitle={Proceedings of the IEEE/CVF Conference on Computer Vision and Pattern Recognition (CVPR)},
  year={2026},
}

@misc{chen2026multimodallifelongunderstandingdataset,
      title={Towards Multimodal Lifelong Understanding: A Dataset and Agentic Baseline}, 
      author={Guo Chen and Lidong Lu and Yicheng Liu and Liangrui Dong and Lidong Zou and Jixin Lv and Zhenquan Li and Xinyi Mao and Baoqi Pei and Shihao Wang and Zhiqi Li and Karan Sapra and Fuxiao Liu and Yin-Dong Zheng and Yifei Huang and Limin Wang and Zhiding Yu and Andrew Tao and Guilin Liu and Tong Lu},
      year={2026},
      eprint={2603.05484},
      archivePrefix={arXiv},
      primaryClass={cs.CV},
      url={https://arxiv.org/abs/2603.05484}, 
}

@inproceedings{alayrac2022flamingovisuallanguagemodel,
      title={Flamingo: a Visual Language Model for Few-Shot Learning}, 
      author={Jean-Baptiste Alayrac and Jeff Donahue and Pauline Luc and Antoine Miech and Iain Barr and Yana Hasson and Karel Lenc and Arthur Mensch and Katie Millican and Malcolm Reynolds and Roman Ring and Eliza Rutherford and Serkan Cabi and Tengda Han and Zhitao Gong and Sina Samangooei and Marianne Monteiro and Jacob Menick and Sebastian Borgeaud and Andrew Brock and Aida Nematzadeh and Sahand Sharifzadeh and Mikolaj Binkowski and Ricardo Barreira and Oriol Vinyals and Andrew Zisserman and Karen Simonyan},
      booktitle={Advances in Neural Information Processing Systems (NeurIPS)},
      year={2022},
}

@inproceedings{li2023blip2bootstrappinglanguageimagepretraining,
      title={BLIP-2: Bootstrapping Language-Image Pre-training with Frozen Image Encoders and Large Language Models}, 
      author={Junnan Li and Dongxu Li and Silvio Savarese and Steven Hoi},
      booktitle={Proceedings of the 40th International Conference on Machine Learning (ICML)},
      year={2023},
}

@inproceedings{tapaswi2016movieqaunderstandingstoriesmovies,
      title={MovieQA: Understanding Stories in Movies through Question-Answering}, 
      author={Makarand Tapaswi and Yukun Zhu and Rainer Stiefelhagen and Antonio Torralba and Raquel Urtasun and Sanja Fidler},
      booktitle={Proceedings of the IEEE Conference on Computer Vision and Pattern Recognition (CVPR)},
      year={2016},
}

@inproceedings{song2023moviechat,
  title={MovieChat: From Dense Token to Sparse Memory for Long Video Understanding},
  author={Song, Enxin and Chai, Wenhao and Wang, Guanhong and Zhang, Yucheng and Zhou, Haoyang and Wu, Feiyang and Guo, Xun and Ye, Tian and Lu, Yan and Hwang, Jenq-Neng and others},
  booktitle={Proceedings of the IEEE/CVF Conference on Computer Vision and Pattern Recognition (CVPR)},
  year={2024}
}

@article{song2024moviechat+,
  title={MovieChat+: Question-aware Sparse Memory for Long Video Question Answering},
  author={Song, Enxin and Chai, Wenhao and Ye, Tian and Hwang, Jenq-Neng and Li, Xi and Wang, Gaoang},
  journal={IEEE Transactions on Pattern Analysis and Machine Intelligence (TPAMI)},
  year={2025},
  pages={374--389},
  doi={10.1109/TPAMI.2025.3604614}
}

@inproceedings{lin-etal-2024-video,
    title = "Video-{LL}a{VA}: Learning United Visual Representation by Alignment Before Projection",
    author = "Lin, Bin  and
      Ye, Yang  and
      Zhu, Bin  and
      Cui, Jiaxi  and
      Ning, Munan  and
      Jin, Peng  and
      Yuan, Li",
    editor = "Al-Onaizan, Yaser  and
      Bansal, Mohit  and
      Chen, Yun-Nung",
    booktitle = "Proceedings of the 2024 Conference on Empirical Methods in Natural Language Processing",
    month = nov,
    year = "2024",
    address = "Miami, Florida, USA",
    publisher = "Association for Computational Linguistics",
    url = "https://aclanthology.org/2024.emnlp-main.342/",
    doi = "10.18653/v1/2024.emnlp-main.342",
    pages = "5971--5984"
}

@inproceedings{maaz2024videochatgptdetailedvideounderstanding,
      title={Video-ChatGPT: Towards Detailed Video Understanding via Large Vision and Language Models}, 
      author={Muhammad Maaz and Hanoona Rasheed and Salman Khan and Fahad Shahbaz Khan},
      booktitle={Proceedings of the 62nd Annual Meeting of the Association for Computational Linguistics (ACL)},
      year={2024},
      pages={12585--12602},
}

@InProceedings{Bain21,
  author       = "Max Bain and Arsha Nagrani and G{\"u}l Varol and Andrew Zisserman",
  title        = "Frozen in Time: A Joint Video and Image Encoder for End-to-End Retrieval",
  booktitle    = "IEEE International Conference on Computer Vision",
  year         = "2021",
}

@inproceedings{lei-etal-2018-tvqa,
    title = "{TVQA}: Localized, Compositional Video Question Answering",
    author = "Lei, Jie  and
      Yu, Licheng  and
      Bansal, Mohit  and
      Berg, Tamara",
    editor = "Riloff, Ellen  and
      Chiang, David  and
      Hockenmaier, Julia  and
      Tsujii, Jun{'}ichi",
    booktitle = "Proceedings of the 2018 Conference on Empirical Methods in Natural Language Processing",
    month = oct # "-" # nov,
    year = "2018",
    address = "Brussels, Belgium",
    publisher = "Association for Computational Linguistics",
    url = "https://aclanthology.org/D18-1167/",
    doi = "10.18653/v1/D18-1167",
    pages = "1369--1379"
}

@misc{videoexplorer2026,
  title={VideoExplorer: Think With Videos For Agentic Long-Video Understanding},
  author={Huaying Yuan and Zheng Liu and Junjie Zhou and Hongjin Qian and Yan Shu and Nicu Sebe and Ji-Rong Wen and Zhicheng Dou},
  year={2025},
  eprint={2506.10821},
  archivePrefix={arXiv},
  primaryClass={cs.CV},
  url={https://arxiv.org/abs/2506.10821}
}

@inproceedings{ding2025streammind,
  title={StreamMind: Unlocking Full Frame Rate Streaming Video Dialogue through Event-Gated Cognition},
  author={Ding, Xin and Wu, Hao and Yang, Yifan and Jiang, Shiqi and Zhang, Qianxi and Bai, Donglin and Chen, Zhibo and Cao, Ting},
  booktitle={Proceedings of the IEEE/CVF International Conference on Computer Vision (ICCV)},
  year={2025},
}

@misc{santos2025inftyvideotrainingfreeapproachlong,
  title={Infinity-Video: A Training-Free Approach to Long Video Understanding via Continuous-Time Memory Consolidation},
  author={Saul Santos and Antonio Farinhas and Daniel C. McNamee and Andre F. T. Martins},
  year={2025},
  eprint={2501.19098},
  archivePrefix={arXiv},
  primaryClass={cs.CV},
  url={https://arxiv.org/abs/2501.19098}
}

@inproceedings{chen2024imageworth12tokens,
  title={An Image is Worth 1/2 Tokens After Layer 2: Plug-and-Play Inference Acceleration for Large Vision-Language Models},
  author={Liang Chen and Haozhe Zhao and Tianyu Liu and Shuai Bai and Junyang Lin and Chang Zhou and Baobao Chang},
  booktitle={European Conference on Computer Vision (ECCV)},
  year={2024},
}

@misc{zou2024from,
  title={From Seconds to Hours: Reviewing MultiModal Large Language Models on Comprehensive Long Video Understanding},
  author={Heqing Zou and Tianze Luo and Guiyang Xie and Victor Zhang and Fengmao Lv and Guangcong Wang and Junyang Chen and Zhuochen Wang and Hansheng Zhang and Huaijian Zhang},
  year={2024},
  eprint={2409.18938},
  archivePrefix={arXiv},
  primaryClass={cs.CV},
  url={https://arxiv.org/abs/2409.18938}
}

@inproceedings{zhang2025flashvstream,
    title={Flash-VStream: Efficient Real-Time Understanding for Long Video Streams}, 
    author={Haoji Zhang and Yiqin Wang and Yansong Tang and Yong Liu and Jiashi Feng and Xiaojie Jin},
    booktitle={Proceedings of the IEEE/CVF International Conference on Computer Vision (ICCV)},
    year={2025},
}

@misc{yang2025streamagentanticipatoryagentsstreaming,
      title={StreamAgent: Towards Anticipatory Agents for Streaming Video Understanding}, 
      author={Haolin Yang and Feilong Tang and Lingxiao Zhao and Xiang An and Ming Hu and Huifa Li and Xinlin Zhuang and Yifan Lu and Xiaofeng Zhang and Abdalla Swikir and Junjun He and Zongyuan Ge and Imran Razzak},
      year={2025},
      eprint={2508.01875},
      archivePrefix={arXiv},
      primaryClass={cs.CV},
      url={https://arxiv.org/abs/2508.01875}, 
}

@misc{arnab2025temporalchainthoughtlongvideo,
      title={Temporal Chain of Thought: Long-Video Understanding by Thinking in Frames}, 
      author={Anurag Arnab and Ahmet Iscen and Mathilde Caron and Alireza Fathi and Cordelia Schmid},
      year={2025},
      eprint={2507.02001},
      archivePrefix={arXiv},
      primaryClass={cs.LG},
      url={https://arxiv.org/abs/2507.02001}, 
}

@misc{pan2025timesearchradaptivetemporalsearch,
      title={TimeSearch-R: Adaptive Temporal Search for Long-Form Video Understanding via Self-Verification Reinforcement Learning}, 
      author={Junwen Pan and Qizhe Zhang and Rui Zhang and Ming Lu and Xin Wan and Yuan Zhang and Chang Liu and Qi She},
      year={2025},
      eprint={2511.05489},
      archivePrefix={arXiv},
      primaryClass={cs.CV},
      url={https://arxiv.org/abs/2511.05489}, 
}

@misc{tstar,
      title={T*: Re-thinking Temporal Search for Long-Form Video Understanding}, 
      author={Jinhui Ye and Zihan Wang and Haosen Sun and Keshigeyan Chandrasegaran and Zane Durante and Cristobal Eyzaguirre and Yonatan Bisk and Juan Carlos Niebles and Ehsan Adeli and Li Fei-Fei and Jiajun Wu and Manling Li},
      year={2025},
      eprint={2504.02259},
      archivePrefix={arXiv},
      primaryClass={cs.CV},
      url={https://arxiv.org/abs/2504.02259}, 
}

@misc{openai2024gpt4o,
  title        = {Hello GPT-4o},
  author       = {{OpenAI}},
  year         = {2024},
  month        = {May},
  howpublished = {\url{https://openai.com/index/hello-gpt-4o/}},
  note         = {Official OpenAI blog post, Accessed: 2026-04-22}
}

@article{wang2024qwen2vl,
  title   = {Qwen2-VL: Enhancing Vision-Language Model's Perception of the World at Any Resolution},
  author  = {Wang, Peng and Bai, Shuai and Tan, Sinan and Wang, Shijie and Fan, Zhihao and Bai, Jinze and Chen, Keqin and Liu, Xuejing and Wang, Jialin and Ge, Wenbin and Fan, Yang and Dang, Kai and Du, Mengfei and Ren, Xuancheng and Men, Rui and Liu, Dayiheng and Zhou, Chang and Zhou, Jingren and Lin, Junyang},
  journal = {arXiv preprint arXiv:2409.12191},
  year    = {2024},
  url     = {https://arxiv.org/abs/2409.12191}
}

@article{chen2024internvl2,
  title   = {InternVL2: Better than the Best---Expanding Performance Boundaries of Open-Source Multimodal Models with the Progressive Scaling Strategy},
  author  = {Chen, Zhe and Wang, Weiyun and Tian, Hao and Ye, Shenglong and Gao, Zhangwei and Cui, Erfei and Tong, Wenwen and Hu, Kongzhi and Luo, Jiapeng and Ma, Zheng and others},
  journal = {arXiv preprint arXiv:2404.16821},
  year    = {2024},
  url     = {https://arxiv.org/abs/2404.16821}
}

@article{li2024llavaonevision,
  title        = {LLaVA-OneVision: Easy Visual Task Transfer},
  author       = {Li, Bo and Zhang, Yuanhan and Guo, Dong and Zhang, Renrui and Li, Feng and Zhang, Hao and Zhang, Kaichen and Li, Yanwei and Liu, Ziwei and Li, Chunyuan},
  journal      = {Transactions on Machine Learning Research},
  year         = {2025}
}

@article{zhang2024llavavideo,
  title        = {LLaVA-Video: Video Instruction Tuning With Synthetic Data},
  author       = {Zhang, Yuanhan and Wu, Jinming and Li, Wei and Li, Bo and Ma, Zejun and Liu, Ziwei and Li, Chunyuan},
  journal      = {Transactions on Machine Learning Research},
  year         = {2025}
}

@inproceedings{videollmonline2024,
  title     = {VideoLLM-online: Online Video Large Language Model for Streaming Video},
  author    = {Chen, Joya and Lv, Zhaoyang and Wu, Shiwei and Lin, Kevin Qinghong and Song, Chenan and Gao, Difei and Liu, Jia-Wei and Gao, Ziteng and Mao, Dongxing and Shou, Mike Zheng},
  booktitle = {Proceedings of the IEEE/CVF Conference on Computer Vision and Pattern Recognition (CVPR)},
  month     = {June},
  year      = {2024},
  pages     = {18407--18418}
}

@inproceedings{qian2025dispider,
  title     = {Dispider: Enabling Video LLMs with Active Real-Time Interaction via Disentangled Perception, Decision, and Reaction},
  author    = {Qian, Rui and Ding, Shuangrui and Dong, Xiaoyi and Zhang, Pan and Zang, Yuhang and Cao, Yuhang and Lin, Dahua and Wang, Jiaqi},
  booktitle = {Proceedings of the IEEE/CVF Conference on Computer Vision and Pattern Recognition (CVPR)},
  year      = {2025},
}

@inproceedings{shen2024longvu,
  title   = {LongVU: Spatiotemporal Adaptive Compression for Long Video-Language Understanding},
  author  = {Shen, Xiaoqian and Xiong, Yunyang and Zhao, Changsheng and Wu, Lemeng and Chen, Jun and Zhu, Chenchen and Liu, Zechun and Xiao, Fanyi and Varadarajan, Balakrishnan and Bordes, Florian and Liu, Zhuang and Xu, Hu and Kim, Hyunwoo J. and Soran, Bilge and Krishnamoorthi, Raghuraman and Elhoseiny, Mohamed and Chandra, Vikas},
  booktitle = {Proceedings of the 42nd International Conference on Machine Learning (ICML)},
  series  = {Proceedings of Machine Learning Research},
  volume  = {267},
  pages   = {54582--54599},
  year    = {2025},
  publisher = {PMLR},
  url     = {https://proceedings.mlr.press/v267/shen25j.html}
}

@misc{bai2025qwen3vltechnicalreport,
      title={Qwen3-VL Technical Report}, 
      author={Shuai Bai and Yuxuan Cai and Ruizhe Chen and Keqin Chen and Xionghui Chen and Zesen Cheng and Lianghao Deng and Wei Ding and Chang Gao and Chunjiang Ge and Wenbin Ge and Zhifang Guo and Qidong Huang and Jie Huang and Fei Huang and Binyuan Hui and Shutong Jiang and Zhaohai Li and Mingsheng Li and Mei Li and Kaixin Li and Zicheng Lin and Junyang Lin and Xuejing Liu and Jiawei Liu and Chenglong Liu and Yang Liu and Dayiheng Liu and Shixuan Liu and Dunjie Lu and Ruilin Luo and Chenxu Lv and Rui Men and Lingchen Meng and Xuancheng Ren and Xingzhang Ren and Sibo Song and Yuchong Sun and Jun Tang and Jianhong Tu and Jianqiang Wan and Peng Wang and Pengfei Wang and Qiuyue Wang and Yuxuan Wang and Tianbao Xie and Yiheng Xu and Haiyang Xu and Jin Xu and Zhibo Yang and Mingkun Yang and Jianxin Yang and An Yang and Bowen Yu and Fei Zhang and Hang Zhang and Xi Zhang and Bo Zheng and Humen Zhong and Jingren Zhou and Fan Zhou and Jing Zhou and Yuanzhi Zhu and Ke Zhu},
      year={2025},
      eprint={2511.21631},
      archivePrefix={arXiv},
      primaryClass={cs.CV},
      url={https://arxiv.org/abs/2511.21631},
      note={Technical report},
}

@misc{qwen3_5_2026,
  title        = {Qwen3.5: Towards Native Multimodal Agents},
  author       = {{Qwen Team, Alibaba}},
  year         = {2026},
  month        = {February},
  howpublished = {\url{https://qwen.ai/blog?id=qwen3.5}},
  note         = {Official blog post, Accessed: 2026-03-21}
}

@misc{gemini_video_understanding_2025,
  title        = {Advancing the Frontier of Video Understanding with Gemini 2.5},
  author       = {{Google Developers}},
  year         = {2025},
  month        = {May},
  howpublished = {\url{https://developers.googleblog.com/gemini-2-5-video-understanding/}},
  note         = {Google Developers Blog, Accessed: 2026-04-21}
}

@misc{gemini3_2025,
  title        = {Gemini 3: Introducing the Latest Gemini AI Model from Google},
  author       = {{Google}},
  year         = {2025},
  month        = {November},
  howpublished = {\url{https://blog.google/products/gemini/gemini-3/}},
  note         = {Official Google blog post, Accessed: 2026-04-21}
}

@inproceedings{wang2024internvideo2,
  title        = {InternVideo2: Scaling Foundation Models for Multimodal Video Understanding},
  author       = {Wang, Yi and Li, Kunchang and Li, Xinhao and Yu, Jiashuo and He, Yinan and Wang, Chenting and Chen, Guo and Pei, Baoqi and Yan, Ziang and Zheng, Rongkun and Xu, Jilan and Wang, Zun and Shi, Yansong and Jiang, Tianxiang and Li, Songze and Zhang, Hongjie and Huang, Yifei and Qiao, Yu and Wang, Yali and Wang, Limin},
  booktitle    = {Computer Vision -- ECCV 2024},
  year         = {2024},
  pages        = {396--416},
  doi          = {10.1007/978-3-031-73013-9_23},
  url          = {https://doi.org/10.1007/978-3-031-73013-9_23}
}

@ARTICLE{4310076,
  author={Otsu, Nobuyuki},
  journal={IEEE Transactions on Systems, Man, and Cybernetics}, 
  title={A Threshold Selection Method from Gray-Level Histograms}, 
  year={1979},
  volume={9},
  number={1},
  pages={62-66},
  doi={10.1109/TSMC.1979.4310076}}

@misc{hafner2023mastering,
  title={ Mastering Diverse Domains through World Models },
  author={ Danijar Hafner and Jurgis Pasukonis and Jimmy Ba and Timothy Lillicrap },
  year={ 2023 },
  eprint={ 2301.04104 },
  archivePrefix={arXiv},
  primaryClass={ cs.AI },
  url={ https://arxiv.org/abs/2301.04104 }
}

@misc{zhu2024is,
  title={ Is Sora a World Simulator? A Comprehensive Survey on General World Models and Beyond },
  author={ Zheng Zhu and Xiaofeng Wang and Wangbo Zhao and Chen Min and Bohan Li and Nianchen Deng and Min Dou and Yuqi Wang and Botian Shi and Kai Wang and Chi Zhang and Yang You and Zhaoxiang Zhang and Zhaoxiang Zhang and Dawei Zhao and Liang Xiao and Jian Zhao and Jiwen Lu and Guan Huang },
  year={ 2024 },
  eprint={ 2405.03520 },
  archivePrefix={arXiv},
  primaryClass={ cs.CV },
  url={ https://arxiv.org/abs/2405.03520 }
}

@misc{team2026openworldlib,
  title={OpenWorldLib: A Unified Codebase and Definition of Advanced World Models},
  author={Team, DataFlow and Zeng, Bohan and Hua, Daili and Zhu, Kaixin and Dai, Yifan and Li, Bozhou and Wang, Yuran and Tong, Chengzhuo and Yang, Yifan and Chang, Mingkun and others},
  year={2026},
  eprint={2604.04707},
  archivePrefix={arXiv},
  url={https://arxiv.org/abs/2604.04707}
}

@misc{nvidia2025cosmos,
  author = {Agarwal, Niket and others},
  title = {Cosmos World Foundation Model Platform for Physical AI},
  year = {2025},
  url = {https://arxiv.org/abs/2501.03575},
  eprint = {2501.03575},
  archivePrefix = {arXiv}
}

@misc{du2024towards,
  author = {Du, Yifan and Zhou, Kun and Huo, Yuqi and Li, Yifan and Zhao, Wayne Xin and Lu, Haoyu and Zhao, Zijia and Wang, Bingning and Chen, Weipeng and Wen, Ji-Rong},
  title = {Towards Event-oriented Long Video Understanding},
  year = {2024},
  url = {https://arxiv.org/abs/2406.14129},
  eprint = {2406.14129},
  archivePrefix = {arXiv}
}

@inproceedings{chen2024longvila,
  author = {Xue, Fuzhao and Chen, Yukang and Li, Dacheng and Hu, Qinghao and Zhu, Ligeng and Li, Xiuyu and Fang, Yunhao and Tang, Haotian and Yang, Shang and Liu, Zhijian and He, Ethan and Yin, Hongxu and Molchanov, Pavlo and Kautz, Jan and Fan, Lijuan and Zhu, Yuke and Lu, Yao and Han, Song},
  title = {LongVILA: Scaling Long-Context Visual Language Models for Long Videos},
  booktitle = {International Conference on Learning Representations (ICLR)},
  year = {2025},
}

@inproceedings{ha2018worldmodels,
  title = {Recurrent World Models Facilitate Policy Evolution},
  author = {Ha, David and Schmidhuber, J{\"u}rgen},
  booktitle = {Advances in Neural Information Processing Systems 31},
  pages = {2451--2463},
  year = {2018},
  publisher = {Curran Associates, Inc.},
  url = {https://papers.nips.cc/paper/7512-recurrent-world-models-facilitate-policy-evolution},
  note = "\url{https://worldmodels.github.io}",
}

@misc{zhang2025videoarm,
  title={VideoARM: Agentic Reasoning over Hierarchical Memory for Long-Form Video Understanding},
  author={Xiaoyi Zhang and Zhaoyang Jia and Zongyu Guo and Jiahao Li and Bin Li and Houqiang Li and Yan Lu},
  year={2025},
  eprint={2512.12360},
  archivePrefix={arXiv},
  primaryClass={cs.CV},
  url={https://arxiv.org/abs/2512.12360}
}

@inproceedings{yeo2026worldmm,
  title={WorldMM: Dynamic Multimodal Memory Agent for Long Video Reasoning},
  author={Woongyeong Yeo and others},
  booktitle={Proceedings of the IEEE/CVF Conference on Computer Vision and Pattern Recognition (CVPR)},
  year={2026},
}

@misc{chen2026telemem,
  title={TeleMem: Building Long-Term and Multimodal Memory for Agentic AI},
  author={Chen, Zhi and Guan, Hao and others},
  year={2026},
  eprint={2601.06037},
  archivePrefix={arXiv},
  primaryClass={cs.AI},
  url={https://arxiv.org/abs/2601.06037}
}

@misc{wang2024lifelongmemory,
  title={LifelongMemory: Leveraging LLMs for Answering Queries in Long-form Egocentric Videos},
  author={Ying Wang and Yanlai Yang and Mengye Ren},
  year={2024},
  eprint={2312.05269},
  archivePrefix={arXiv},
  primaryClass={cs.CV},
  url={https://arxiv.org/abs/2312.05269}
}
